\documentclass[journal]{IEEEtran}
\usepackage{amsmath,amsfonts}
\usepackage{algorithmic}
\usepackage{algorithm}
\usepackage{array}
\usepackage[caption=false,font=normalsize,labelfont=sf,textfont=sf]{subfig}
\usepackage{textcomp}
\usepackage{stfloats}

\usepackage{url}
\usepackage{xcolor}
\usepackage{verbatim}
\usepackage{graphicx}
\usepackage{cite}
\newcommand{\eg}{\textit{e.g.}}
\newcommand{\ie}{\textit{i.e.}}
\newcommand{\etal}{\textit{et al.}}
\usepackage[utf8]{inputenc}
\usepackage{threeparttable}
\usepackage{booktabs}
\usepackage{multirow}
\usepackage{array}
\usepackage{hyperref}
\usepackage{tabularx}
\usepackage{mdwmath}
\usepackage{mdwtab}
\usepackage{eqparbox}
\usepackage{url}
\usepackage{soul}
\usepackage[normalem]{ulem}



\title{ClassLIE: Structure- and Illumination-Adaptive Classification for Low-Light Image Enhancement}

\author{Zixiang Wei$^*$, Yiting Wang$^*$, Lichao Sun$^\dagger$,~\IEEEmembership{IEEE Member}, Athanasios V. Vasilakos,~\IEEEmembership{IEEE Senior Member},\\ Lin Wang$^\dagger$,~\IEEEmembership{IEEE Member}

\thanks{$^*$ These two authors contributed equally to this work. The work was done during internship at VLIS LAB. $^\dagger$ Corresponding authors}
\thanks{Z. Wei and Y. Wang are with National Automotive Innovation Centre, IV sensor group, WMG, The University of Warwick Email: zixiang.wei@warwick.ac.uk, yiting.wang.1@warwick.ac.uk .}
\thanks{L. Sun is with Dept. of Computer Science and Engineering, Lehigh University. Email: lis221@lehigh.edu}
\thanks{Athanasios V. Vasilakos is with the Center for AI Research (CAIR), University of Agder(UiA), Grimstad, Norway. Email: thanos.vasilakos@uia.no}
\thanks{L. Wang is with the VLIS LAB of Artificial Intelligence Thrust, HKUST-GZ, and Dept. of Computer Science and Engineering, HKUST, Hong Kong SAR, China. E-mail: linwang@ust.hk} 
}

\begin{document}
\maketitle
\begin{abstract}
Low-light images often suffer from limited visibility and multiple types of degradation, rendering low-light image enhancement (LIE) a non-trivial task. Some endeavors have been recently made to enhance low-light images using convolutional neural networks (CNNs). However, they have low efficiency in learning the structural information and diverse illumination levels at the local regions of an image. Consequently, the enhanced results are affected by unexpected artifacts, such as unbalanced exposure, blur, and color bias. To this end, this paper proposes a novel framework, called ClassLIE, that combines the potential of CNNs and transformers. It classifies and adaptively learns the structural and illumination information from the low-light images in a holistic and regional manner, thus showing better enhancement performance. 
Our framework first employs a structure and illumination classification (SIC) module to learn the degradation information adaptively. In SIC, we decompose an input image into an illumination map and a reflectance map. A class prediction block is then designed to classify the degradation information by calculating the structure similarity scores on the reflectance map and mean square error on the illumination map. As such, each input image can be divided into patches with three enhancement difficulty levels. Then, a feature learning and fusion (FLF) module is proposed to adaptively learn the feature information with CNNs for different enhancement difficulty levels while learning the long-range dependencies for the patches in a holistic manner. Experiments on five benchmark datasets consistently show our ClassLIE achieves new state-of-the-art performance, with 25.74 PSNR and 0.92 SSIM on the LOL dataset.
\end{abstract}




\vspace{-5pt}
\section{Introduction}

Images taken under sub-optimal illumination conditions are affected by different types of degradation, \eg,~poor visibility, low contrast, increased noise, and inaccurate color rendering~\cite{li2021low}. Therefore, low-light image enhancement (LIE) is a crucial task that aims to improve low-light image quality. However, due to the highly ill-posed property of LIE, it remains a challenging low-level vision task.

\begin{figure}[t!]
 \begin{center}
 \begin{minipage}[t]{0.32\linewidth}
    \centerline{\includegraphics[width=\textwidth]{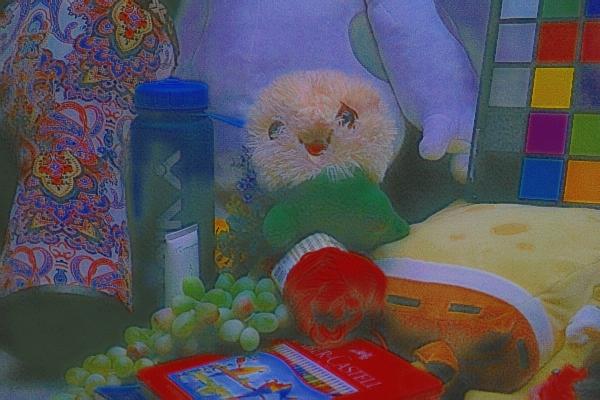}}
    \vspace{0pt}
    \centerline{\small RetinexNet}
     \vspace{3pt}
\centerline{\includegraphics[width=\textwidth]{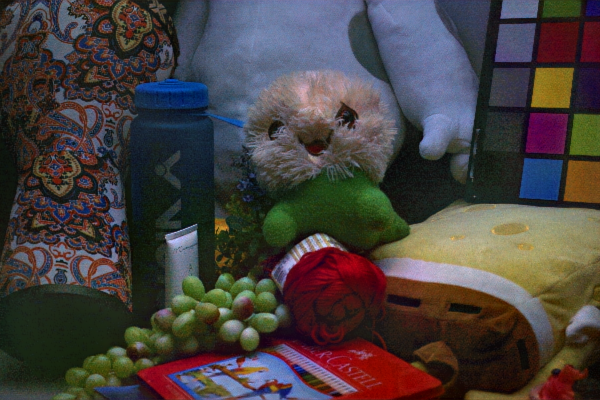}}
    \centerline{\small EnlightenGAN}
 \end{minipage}
 \begin{minipage}[t]{0.32\linewidth}
  \centerline{\includegraphics[width=\textwidth]{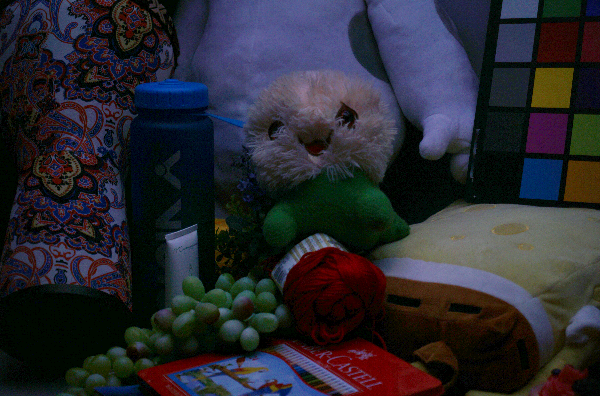}}
  \vspace{0pt}
  \centerline{\small Zero-DCE++}
  \vspace{3pt}
  \centerline{\includegraphics[width=\textwidth]{exp-LOL/zerodce++493.png}}
  \centerline{\small KinD++}  
 \end{minipage}
 \begin{minipage}[t]{0.32\linewidth}  \centerline{\includegraphics[width=\textwidth]{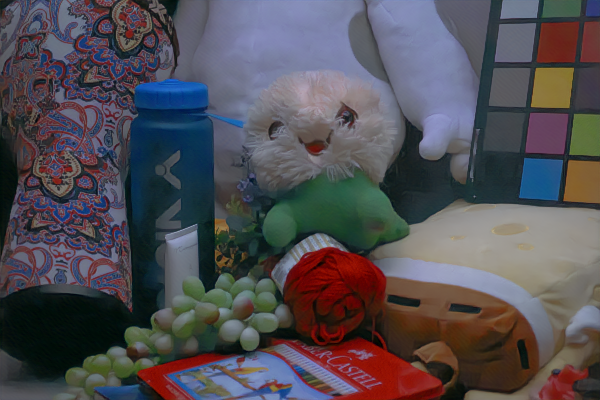}}
  \centerline{\small URetinex-Net}
  \vspace{3pt} 
  \centerline{\includegraphics[width=\textwidth]{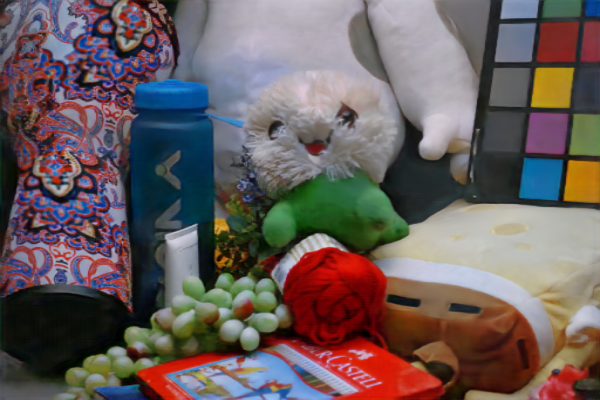}}
  \centerline{\small Ours}  
 \end{minipage}
 \vspace{3pt}
 \caption{Visual comparison with SoTA methods on the LOL V1 dataset. Our method achieves better results with less noise and better exposure, structural and color details.}
 \label{Intro}
\end{center}
\vspace{-20pt}
\end{figure}

Recently, with the development of deep learning, the performance of LIE has been remarkably boosted, and abundant works have been proposed based on convolutional neural networks (CNNs)~\cite{lore2017llnet,wei2018deep,zhang2019kindling,zhu2020eemefn}. Early methods, \eg,~\cite{lore2017llnet,wei2018deep,liu2021retinex}, enhance the low-light images by learning holistic image information based on the Retinex theory~\cite{land1977retinex}; however, they ignore the local structure and illumination information of the low-light images, resulting in visible noise and blur in the enhanced results. For this reason, KinD Net \cite{zhang2019kindling} tackles the noise distribution by using the decomposed illumination map as guidance for LIE. Moreover, some methods explore additional prior information, \eg,~semantic information~\cite{zheng2022semantic,liang2021semantically}, color map detection~\cite{wang2022low} or edge detection \cite{zhu2020eemefn}, for guiding the LIE networks. However, these methods suffer from two critical limitations: 1) the discrepancy of the illumination and structure at the different local regions has been ignored as the pixel regions in an image are treated equally. Consequently, unexpected artifacts appear in the enhanced results, such as unbalanced exposure and blur; 2) color distortion often occurs in the over-/under-exposed regions of the enhanced results.
\vspace{-1pt}

In this paper, \textit{we cast LIE to a classification problem where we strive to classify the enhancement difficulty levels of the low-light images based on illumination and structure prior information and adaptively learn the feature in a holistic and region-wise manner}. To this end, we propose a novel framework, called \textbf{ClassLIE}, that subtly combines the potential of CNNs and transformers~\cite{vaswani2017attention} to enhance the low-light images through class-adaptive feature learning. Overall, our method yields visually better results, with less noise or under-/over- exposure problems and better structure details, as shown in Fig.~\ref{Intro}. Our ClassLIE framework consists of two important components: the structure and illumination classification (SIC) module and the feature learning and fusion (FLF) module. The SIC module decomposes a low-light image into an illumination map and a reflectance map. Then, a class prediction block is designed to classify input patches into three enhancement difficulty levels (\ie~easy, medium, and hard). This is achieved by calculating the structure similarity scores on the reflectance map and mean square error on the illumination map. Accordingly, the enhancement difficulty levels for each patch can be predicted. Then, the FLF module learns the feature information with CNNs for different enhancement difficulty levels in a class-adaptive manner, while learning the long-range dependencies for the patches in a holistic manner. The proposed FLF module consists of a class-adaptive feature learning branch (CAFL) to adaptively learn the class-wise patch features and a patch relation learning (PRL) branch to capture the long-range dependencies. Finally, as the reflectance maps contain correct color information that is invariant to the illumination, we add global residual learning to the decomposed reflectance to ensure color consistency.

We conduct extensive experiments on five benchmark datasets: LOL~\cite{wei2018deep}, LIME~\cite{guo2016lime}, MEF~ \cite{ma2015perceptual}, NPE~\cite{wang2013naturalness} and DICM~\cite{lee2013contrast}. The experimental results show that our method consistently surpasses the previous methods, \eg,~\cite{wu2022uretinex} by a large margin.

In summary, our contributions are four folds: (\textbf{I}) We are the first to cast LIE to a classification problem and propose the ClassLIE framework for LIE; (\textbf{II}) we introduce the SIC module to classify both the structure and illumination information from the decomposed reflectance and illumination maps; (\textbf{III}) we propose the feature learning and fusion module subtly combines the potential of CNNs and transformer to learn the holistic and class-adaptive feature information, resulting in more naturally enhanced results.
\vspace{-3pt}
\section{Related Work}
\subsection{Low-Light Image Enhancement}
Traditional methods for LIE are mainly based on histogram equalization (HE) \cite{pizer1990contrast} and Retinex theory \cite{land1977retinex}. Specifically, HE-based methods enhance images by remapping the intensity levels based on the distribution while Retinex-based methods enhance the low-light images based on the decomposed reflectance and illumination maps. For example, Guo \etal~\cite{guo2016lime} aimed to enhance the image by estimating its illumination map, optimized by a structure prior. However, the handcrafted scheme is not adaptive enough, and they ignore the spatial information of the images. 

LLNet \cite{lore2017llnet} is the first supervised approach using an encoder-decoder architecture. RetinexNet \cite{wei2018deep} decomposes a low-light image into illumination and reflectance layers and enhances low-light images by adjusting the illumination and reconstructing the reflectance. Similarly, KinD~\cite{zhang2019kindling} uses the Retinex theory and three subnets. URetinex-Net \cite{wu2022uretinex} proposes the deep unfolding network to decompose the low-light image into illumination and reflectance. To tackle the data-hungry problem, unsupervised learning \cite{jiang2021enlightengan,ma2022toward,liu2021retinex,wang2021deep}, zero-shot learning \cite{guo2020zero} and semi-supervised learning \cite{yang2020fidelity} approaches have been proposed. However, without enough labeled data for supervision, the details and noise distribution of the low-light images can not be fully restored~\cite{yang2020fidelity}. 

Consequently, some methods concentrate on improving the illumination details or structural texture information on enhancement results.  Illumination maps are used in the frameworks of KinD~\cite{zhang2019kindling} and EnlightenGAN\cite{jiang2021enlightengan} to guide illumination restoration. On the other hand, DCCNet~\cite{zhang2022deep} introduces a predicted gray image during the enhancement process to guide the network's enhancement effects. Similarly, LLFlow~\cite{wang2022low} employs a color map to ensure color consistency between the enhanced low-light images and the original images. SMG~\cite{SMG} also utilizes structure guidance to direct the recovery of edge information. Our method is essentially supervised, and we regard LIE as a classification problem where we utilize both illumination and structure prior information and adaptively enhance the feature information in a holistic and region-wise manner. 

\begin{figure*}[t!]
\centering
\includegraphics[width=.96\textwidth]{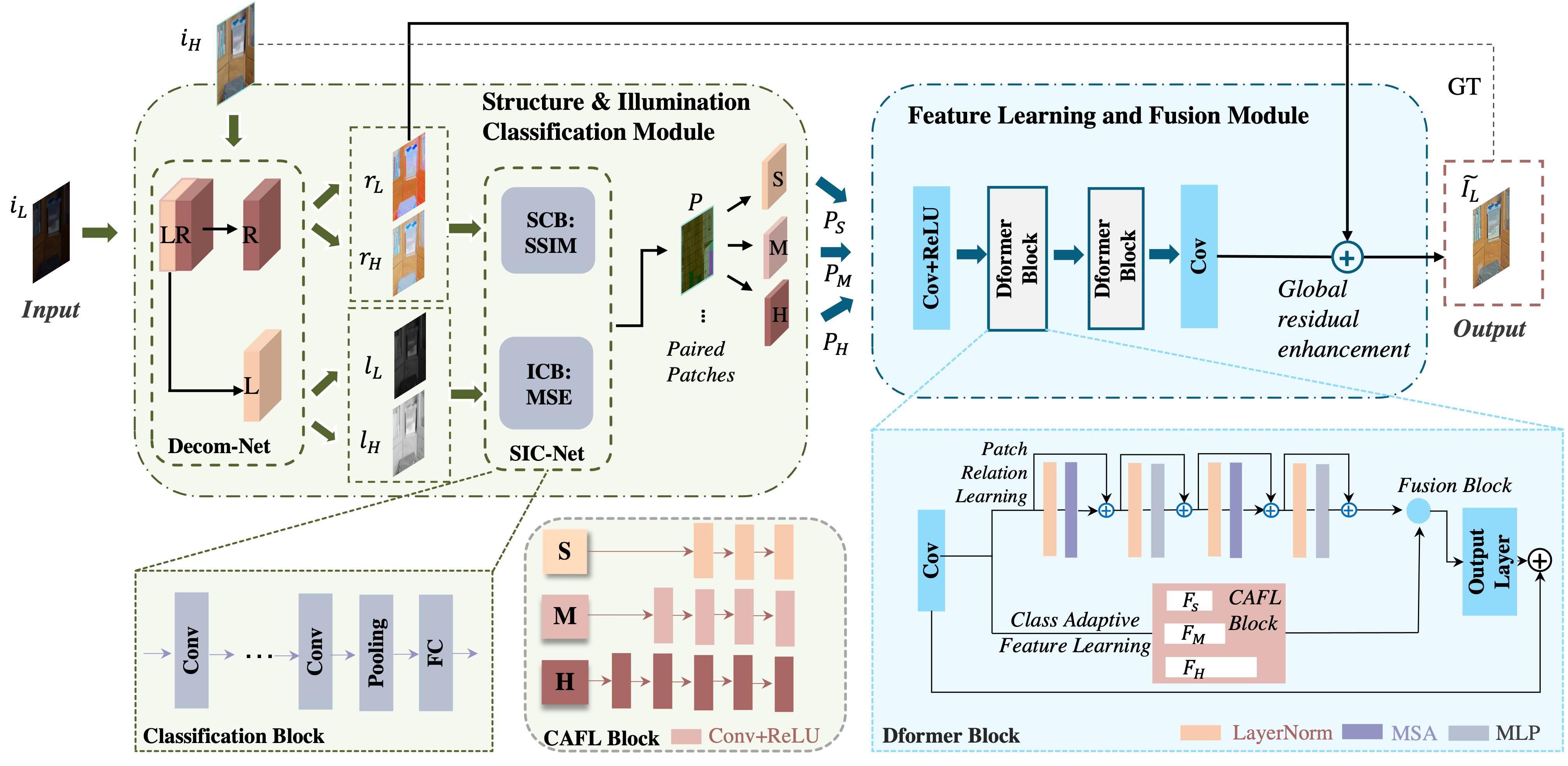}
\caption{Illustration of our ClassLIE framework. Details are shown in the main contexts. It consists of a structure and illumination classification (SIC) module and a feature learning and fusion (FLF) module. Firstly, the Decom-Net decomposites the input low-light images into reflectance ($R$) and illumination ($L$). Then, the SIC net which consists of the structural and illumination classification blocks trains the classification model using the paired low-light and normal-light patches $P$ to classify a random patch into three different enhancement difficulty levels (i.e. S, M, H). After that, with the classified image patches (\ie, $P_{S}, P_{M}, P_{H}$) as the input for the FLF module, the final enhanced result $\tilde{I_{L}}$ can be obtained.}
\vspace{-12pt}
\label{method}
\end{figure*}

\subsection{Vision Transformers}
The transformer \cite{vaswani2017attention} has shown to be effective for natural language processing. 
Recently, the transformer has been shown effective for high-level vision tasks~\cite{liu2021swin,fan2022sunet,chu2021twins,liu2021swin} while some research has demonstrated the potential of transformer for low-level vision tasks \cite{yang2020learning,chen2021pre,wang2021uformer,fan2022sunet}. TTSR \cite{yang2020learning} uses both hard and soft attention for image super-resolution. IPT~\cite{chen2021pre} proposes a pre-trained model with a shared transformer body for learning multiple tasks. However, its performance highly depends on large-scale data and multitask correlations. Therefore, Uformer \cite{wang2021uformer} introduces a U-shaped transformer for image restoration in a more efficient and effective way. SwinIR \cite{liang2021swinir} updates Swim Transformer~\cite{liu2021swin} and adapts it to the image restoration task. Transformer-GAN \cite{yang2022rethinking} combines adversarial learning and transformer in an unsupervised manner.
The recent work \cite{xu2022snr} designs an attention map to guide two-branch feature learning for regions that require long-range or short-range operation based on their different SNR values. However, the structural information is not considered in their design. Our FLF module is built on SwinIR and CNNs. However, it shares a different spirit as it adaptively enhances the classified feature information reflecting the enhancement difficulty levels. Accordingly, we propose the CAFL branch and add it to the PRL branch to achieve the goal of LIE based on structure and illumination classification (See Fig.~\ref{method}).      

\section{Methodology}
\noindent\textbf{Problem Statement.} An overview of the proposed ClassLIE framework is depicted in Fig.~\ref{method}. Denote $i_{L}$ as the low-light image and $i_{H}$ as the corresponding normal-light image, \ie, the ground truth (GT). Generally, we aim to learn a mapping function $F$ that maps an $i_{L}$ to get an enhanced image $\tilde{I_{L}} = F(i_{L})$. To this end, we classify the low-light image patches according to their enhancement difficulty levels and adaptively learn the feature information. By leveraging the class prediction results from SIC module, we design the Dformer block which combines the advantages of both the CNN and transformer to flexibly learn the global and regional features. Specifically, the patch relation learning branch with the transformer architecture helps our model to learn long-range dependencies; while the class-adaptive feature learning branch uses three different CNNs (\ie, $F_{S}, F_{M}, F_{H}$) to learn the classified patch features, which can preserve better regional detail information for enhancement. 
The feature fusion block is designed here to fuse the two features from the two branches with the learned channel dependencies. After the global residual enhancement, the final enhanced result can be obtained. 
\vspace{-5pt}
\subsection{Structure and Illumination Classification (SIC)}
The SIC module determines ``whether the input image patches are easy or hard to enhance" through the structure and illumination information. As shown in Fig.~\ref{method}, our SIC module consists of two steps: layer decomposition and class prediction. 

\noindent\textbf{Step 1 Layer Decomposition.}~~
 Since the illumination of the input images is low and the structural information is fuzzy, we first decompose a low-light image into an illumination map and a reflectance map, according to Retinex theory~\cite{liu2021retinex}. For convenience, we use RetinexNet~\cite{wei2018deep} for layer decomposition without the hand-crafted functions. As shown in Fig.\ref{method}, our ClassLIE takes the paired low-light images $i_{L}$ and normal-light images $i_{H}$ as inputs, and obtains the reflectance maps $r_{L}$, $r_{H}$ and illumination maps $l_{L}$, $l_{H}$, respectively. 
 The decomposed results are used for enhancement difficulty classification in step 2.

\noindent\textbf{Step 2 Class Prediction.}~~
We now classify the enhancement difficulty levels based on the patch structure and illumination maps obtained in Step 1. The structural and illumination information are applied as classification criteria because they are core factors influencing the effect of LIE. Therefore, we design a SIC net that consists of a structure classification block (SCB) and an illumination classification block (ICB). The SCB estimates the structural complexity by measuring the structural similarity (SSIM)~\cite{wang2004image} on the reflectance map as the reflectance delivers rich structure information, similar to that of the normal-light image. The ICB estimates the illumination level of each patch by measuring the mean square error (MSE) on the illumination map, which is reasonable as it removes unnecessary image components. 
 
Our SIC net classifies the patches of the reflectance map and illumination map into three classes: `simple', `medium', and `hard', depending on their enhancement difficulty. The rationale behind the three division levels is grounded in the theoretical foundation from \cite{simonson1948effects}, which explored the impact of three levels of illumination (5 versus 100, 5 versus 300, and 100 versus 300 foot candles) on human perception. This tripartite classification has been subsequently employed in visual studies, as evident in literature \cite{kukula2004evaluation,laddi2013classification}, to categorize illumination conditions into three groups: bright, less bright, and crossover areas. We now explain the training process and test process for class prediction.
 
\begin{figure*}[t!]
\centering
\includegraphics[width=1\textwidth]{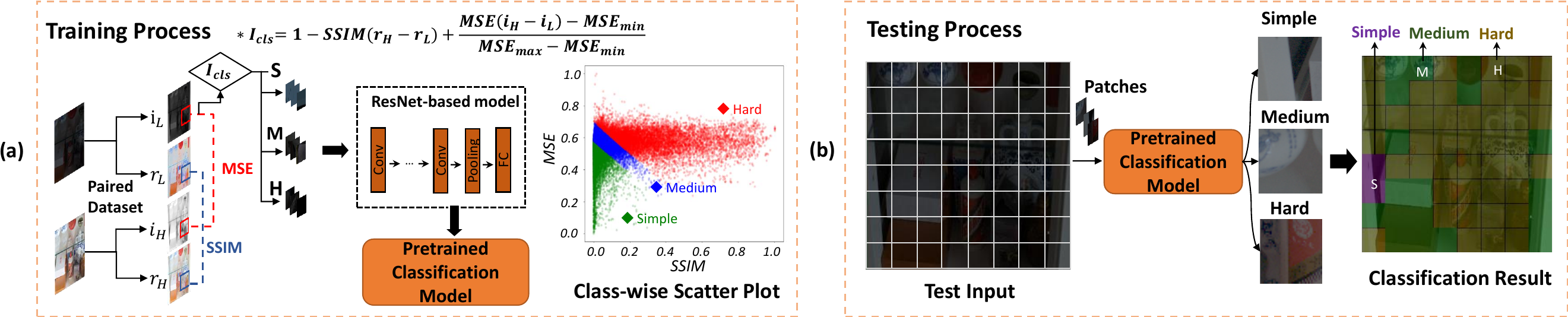}
\caption{Illustration for the train and test processes of our SIC module. (a) The class-wise scatter plot depicts a plot of class distribution with respect to the metrics (SSIM and MSE) for all patches in the LOL training  set. (b) Classification results show the visual patch categories on a low-light example test image after classification. It's important to note that the pre-trained Layer Decomposition network is not involved in the testing process but only used during training in SIC.}
\vspace{-5pt}
\label{class}
\end{figure*}

\noindent\textbf{Training process.}~~ The class labels are generated via the class prediction index in the training process, as shown in Fig.~\ref{class}(a). To determine the separatrix of categories, we calculate the class prediction index $I_{cls}$ on each patch, which is formulated as: 
\vspace{-5pt}
\begin{equation}
\mathcal{I}_{cls} = 1- {ssim(r_{H}-r_{L})+ \frac{mse(i_{H}-i_{L})-mse_{min}}{mse_{max}-mse_{min}}},
\label{eqN9}
\end{equation}

where ${ssim}(\cdot,\cdot)$ denotes the structural difference between a normal-light patch $r_{H}$ and a low-light patch $r_{L}$. ${mse}(\cdot,\cdot)$ measures the illumination MSE difference between a normal-light reflectance patch $i_{H}$ and a low-light reflectance patch $i_{L}$, while $mse_{max}$ and $mse_{min}$ denotes the maximum and minimum illumination MSE values within all the patches. The value of the index is mapped into a range of $[0,1]$, making it convenient to divide the whole data set into numerical gradients. 

After sorting all the patches in the training data according to the index value from the small to the large, we divide them into three equal groups in quantity, corresponding to the categories: `simple', `medium', and `hard', respectively. For instance, we select the index trisections of 0 to 1 (\ie [0, 0.277], (0.277,0.51], (0.51,1]) to divide the LOL training dataset~\cite{wei2018deep} into three categories with the same number of patches. It can be observed that the larger index value of the patch indicates a more complex structure and worse illumination conditions.

To showcase the effectiveness of this classification, we have included a class-wise scatter plot in Fig.~\ref{class}, illustrating that when utilizing three difficulty levels, each patch is accurately classified into its corresponding difficulty category. As shown from the Fig.~\ref{class}, during training, the sorted image patches are fed to a classification network containing five convolution layers, an average pooling layer and a fully connected layer to obtain the probability vector. Inspired by ~\cite{kong2021classsr}, we introduce a classification loss $L_{c}$ and a selection loss $L_{s}$. Specifically, $L_{c}$ loss is the negative number of distance sum between each class probability for the same patch. 
It ensures that the classification network has higher confidence and lower entropy, formulated as:
\begin{equation}
\centering
L_{c} = -\sum_{i = 1}^{M-1} \sum_{j = i+1}^{M}\left|P_{i}(x)-P_{j}(x)\right|, \text { s.t. } \sum_{i = 1}^{M} P_{i}(x) = 1 .
\label{eqN1}
\end{equation}
where $M$ is the number of classes. Selection loss $L_{s}$ aims to divide each patch with an equal probability into three categories. It makes the mean value of multiple predictions more dispersed in order that three categories are involved with the same probability. 
It can be formulated as:
\begin{equation}
\centering
L_{s}=\sum_{i=1}^{M}\lvert\sum_{j=1}^{B} P_{i}\left(x_{j}\right)-\frac{B}{M}\rvert,
\label{eqN2}
\end{equation}
where $B$ is the batch size, $L_{s}$ is the sum of the distance between the average number $\frac{B}{M}$ and the number of patches for each class in the batch. To propagate gradients, we use the probability sum $\sum_{j=1}^{B} P_{i}$ to calculate the patches number rather than the statistic number. In addition, the commonly used $L_{1}$ loss is also used for the image loss function. Therefore, the total classification loss $\mathcal{L}_{cls}$ is the linear combination of $L_{1}$, $L_{c}$ and $L_{s}$, formulated as:
\vspace{-2pt}

\begin{equation}
\mathcal{L}_{cls} = w_{1} \times L_{1} + w_{2} \times L_{c} + w_{3} \times L_{s},
\label{eqN3}
\end{equation}

where $w_{1}$,$w_{2}$ and $w_{3}$ are the balancing weights. 

\noindent\textbf{Test Process.} As shown in Fig.~\ref{class}~(b), In terms of the testing phase, each test RGB input is divided into 64 patches of size 32 × 32 for input. These patches are then learned and classified into the `simple', `medium', and `hard' categories by the pre-trained SIC classification model. The SIC pre-trained model used for testing has already retained the structure and weights of the classification network, allowing it to accurately classify the patches of a test set image without ground truth into specific enhancement difficulty levels.

It's significant to underscore that the RetinexNet~\cite{wei2018deep} Layer Decomposition network, to which we alluded, need not be incorporated in this. This is attributable to the pre-trained SIC network's successful learning of the feature distribution of `simple', `medium', and `hard' patch categories, thereby negating the necessity for decoupling computations during the testing phase. The Layer Decomposition network is only used for the computation of structural and illumination components in the training of the SIC module.
\vspace{-5pt}
\subsection{Feature Learning and Fusion (FLF)}
The FLF module is designed to simultaneously learn the patch relation information and the class-wise feature information according to the three enhancement difficulty levels. 
As shown in Fig.~\ref{method}, we use a $3 \times 3$ convolutional layer with the ReLU as the activation function to extract features from the low-light patches. Then, two Dformer blocks are designed to enhance the classified patch features, which are fused with the global residual learning, to get an enhanced image. 
We now explain the details of the Dformer block and global residual learning.

\noindent\textbf{Dformer Block.}~
The Dformer block is a novel and advanced structure that harmonizes the benefits of both transformer networks and convolutional neural networks (CNNs). This innovative model is structured with two main branches, namely, a patch relation learning (PRL) branch and a class-adaptive feature learning (CAFL) branch. These branches work in tandem to deliver a robust and efficient output. The first branch PRL is responsible for extracting and learning relationships among different patches of the input. This is achieved through the application of Swin transformer layers, a choice influenced by their proven effectiveness in low-level vision tasks. The second branch, CAFL is designed to adaptively learn the feature information based on the complexity of the enhancement process.

Adding to the architecture's complexity, an additional $3 \times 3$ convolutional layer is incorporated into the model, contributing to the extraction of more intricate and detailed features from the input.  On top of all these components, the Dformer employs global residual learning. The primary function of this mechanism is to ensure the preservation of color consistency across the image to ensure better metrics and more natural visual results.

    
\noindent\textbf{Patch Relation Learning (PRL)}~Considering the limitations of CNNs in modeling the long-range dependencies, we propose PRL to obtain more sufficient holistic contextual information. In our work, we use the Swin transformer layers \cite{liu2021swin} as the basic layers because it has been proven effective in low-level vision tasks. We use two transformer layers to balance performance and parameters. With PRL, we get the enhanced patch feature $F_{p^{i}}$ after the $i$-th Dformer block.

\noindent\textbf{Class-adaptive Feature Learning (CAFL)}~CAFL aims to adaptively learn the feature information with respect to the enhancement difficulty levels. For this, we design a CAFL block consisting of three CNNs $F_{S}$, $F_{M}$ and $F_{H} $, each of which has a different number of layers (See Fig.~\ref{method}). In our work, we use DnCNN~\cite{zhang2017beyond} as our backbones as it is shown to be more effective than the commonly used U-Net ~\cite{ronneberger2015u} (See Tab.~\ref{baseline}). The reason is that U-Net requires a relatively high feature dimension; thus, the receptive field covered by a single pixel is large. By contrast, DnCNN does not involve dimensional upsampling or downsampling. Therefore, it can avoid the shadow issues, caused by the high receptive field. We remove the batch normalization of DnCNN because it shows a negative impact on the enhancement results in the ablation study (See suppl. material). In terms of the `simple' category, we use three combination layers of $ 3 \times 3$ convolution while for the `medium' and `hard' categories, the number of layers is increased to four and five, respectively. Consequently, for $i$-th Dformer block, we can get the class-adaptive features $F_{c} $ from the classified patch features (\ie, $P_{S^{i}}, P_{M^{i}}, P_{H^{i}}$):
    \begin{equation}
     F_{c^{i}} = F_{S}(P_{S^{i}}) \oplus F_{M}(P_{M^{i}}) \oplus F_{H}(P_{H^{i}}),  
    \label{eqN4}
    \end{equation}
    where the $\oplus $ denotes the concatenation operation.
Eventually, we can obtain the CAFL feature $F_{c^{i}}$ after the $i$-th Dformer block.







\begin{figure}[t]
\centerline{\includegraphics[width=\columnwidth]{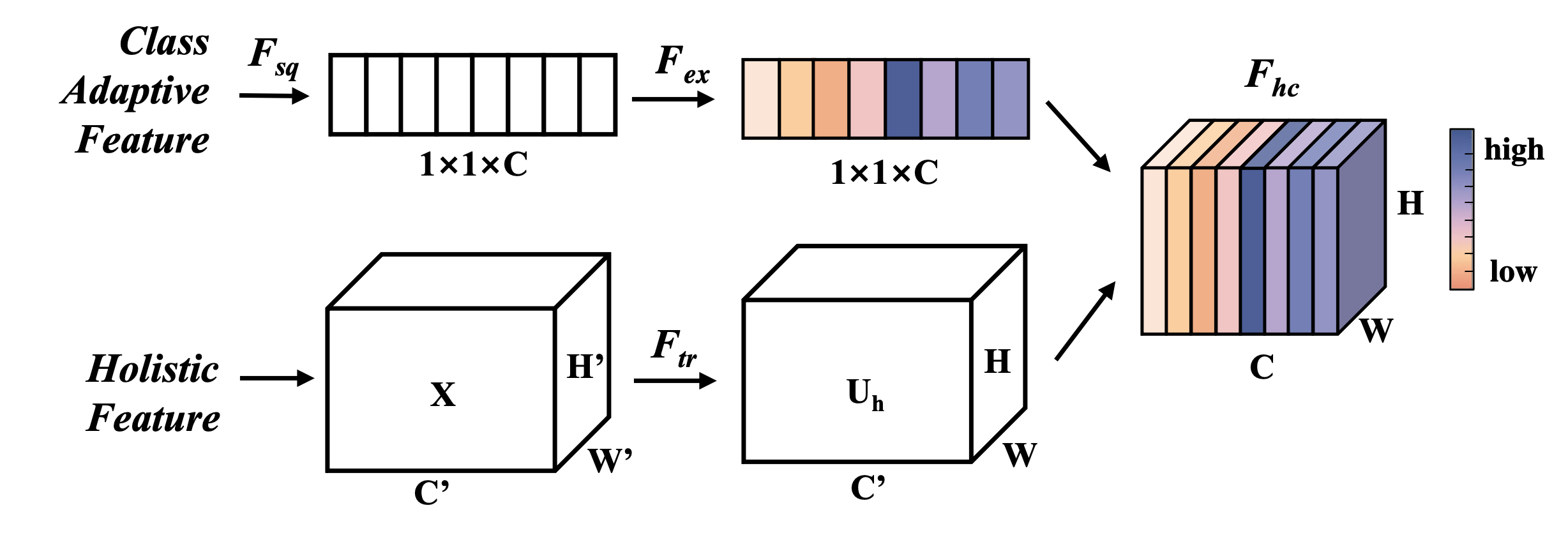}}

\caption{An illustration of feature fusion block. Note that darker color (purple) illustrates higher attention weights, while lighter color (yellow) illustrates higher attention weights. }
\vspace{-12pt}
\label{BL}
\end{figure}

\noindent\textbf{Feature Fusion Block.}~
With the features $F_{p^{i}}$ from PRL branch and $F_{c^{i}}$ from the CAFL branch, we now fuse both types of features (See Fig.~\ref{method}). Since the two branches take image patches as inputs, boundary artifacts, and information loss are introduced for each image patch. Although this problem can be tackled with patch overlap~\cite{liang2021swinir}, it cannot handle the arbitrary input feature and it leads to considerable computation costs. To tackle this problem, we propose a fusion block via channel attention, inspired by SENet~\cite{hu2018squeeze}. As the channel attention mechanism is shown to be effective in modeling the interdependencies between channels by learning the feature hierarchy, we leverage the class-adaptive features to guide the holistic patch relation features to model the dependencies between the two feature channels.  we use two feature maps as input while they only use one feature for this operation. A diagram illustrating the structure of our feature fusion block is shown in Fig.~\ref{BL}. 
In our fusion structure, firstly, the two features go through a transformation $ F_{tr}$ mapping to get the characteristics with the calculation of:
\begin{equation}
    U_{p} = F_{tr}(F_{p^{i}})
    \label{euq5}
\end{equation}
\begin{equation}
    U_{c} = F_{tr}(F_{c^{i}})
    \label{euq6}
\end{equation}
where the feature size is changes from $ \in R^{H^{\prime} \times W^{\prime} \times C^{\prime}}$ into $ \in R^{H \times W \times C}$. Let $V = [{v}_{1},{v}_{2},...,{v}_{c}]$ denote the learned set of filter kernels, where ${v}_{c}$ refers to the parameters of the c-th filter, and the output for the class adaptive feature is $U_{c} = [{u}_{c1},{u}_{c2},...,{u}_{cc}]$. This is calculated with 

\begin{equation}
    u_{cc}=v_{c} * X=\sum_{s=1}^{C^{\prime}} v_{c}^{s} * F_{c}^{s}
    \label{euqa7}
\end{equation}
where the $F_{c}^{s}$ represents the s-th input of the class-adaptive feature. It is also the same operation for the $U_{p}$, where the $u_{pc}$ denotes the feature learned with the filter kernels. 

In our feature fusion structure, we follow the SENet to do squeeze and excitation, differently, we use two feature maps as input while they only use one feature for this operation.

\noindent\textbf{Global Residual Enhancement.}~Finally, after we get the initial enhanced low-light images $\bar{i_{L}}$ after all Dformer blocks. We combine $\bar{i_{L}}$ with the initial low-light reflectance $r_{L}$ that contains rich color and structure information to restore the color information, where we can get our final enhanced result $\tilde{I_{L}}$ by
\begin{equation}
\tilde{I_{L}} =  \bar{i_{L}} \oplus r_{L}
\label{eqN8}
\end{equation}

We use L1 loss between the enhanced result $ \tilde{I_{L}}$ and the GT normal-light image $i_{H}$ by  $ \mathcal{L}_{enhance} = \parallel \tilde{I_{L}} - i_{H}\parallel _{1} $.

\begin{figure*}[t]
 \begin{center}
 \begin{minipage}[t]{0.137\linewidth}
    \centerline{\includegraphics[width=\textwidth]{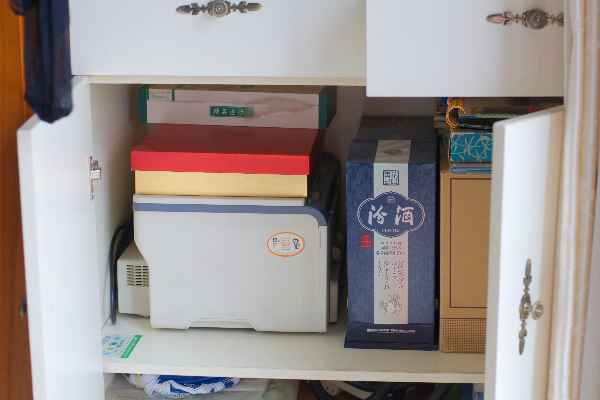}}
    \vspace{1.5pt}
    \centerline{\includegraphics[width=\textwidth]{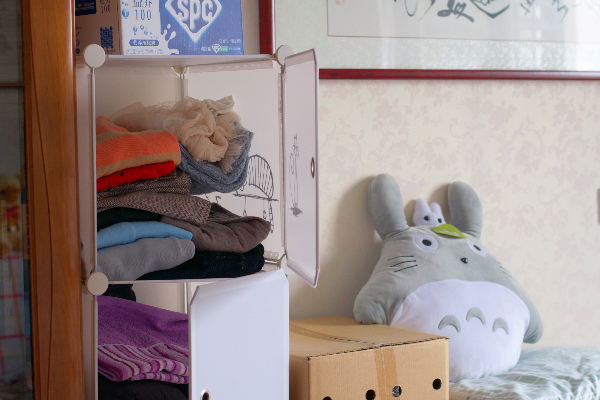}}
    \vspace{1.5pt}
    \centerline{\includegraphics[width=\textwidth]{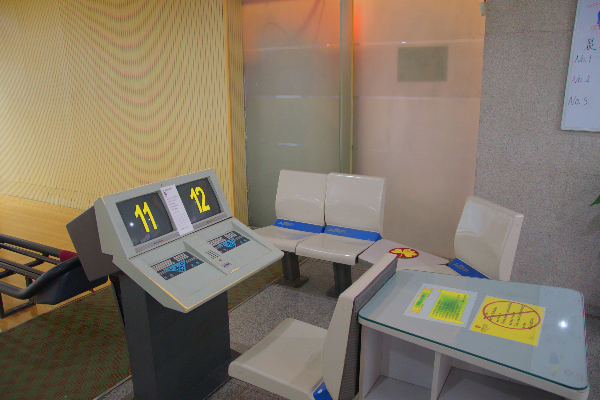}}
    \vspace{1.5pt}
    \centerline{GT}
   \end{minipage}
 \begin{minipage}[t]{0.137\linewidth}
   \centerline{\includegraphics[width=\textwidth]{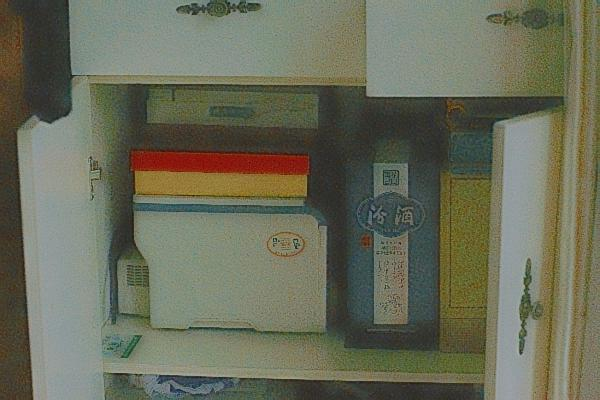}}
   \vspace{1.5pt}
   \centerline{\includegraphics[width=\textwidth]{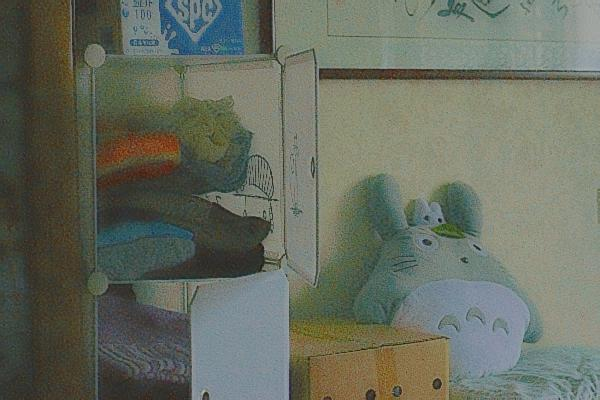}}
   \vspace{1.5pt}
   \centerline{\includegraphics[width=\textwidth]{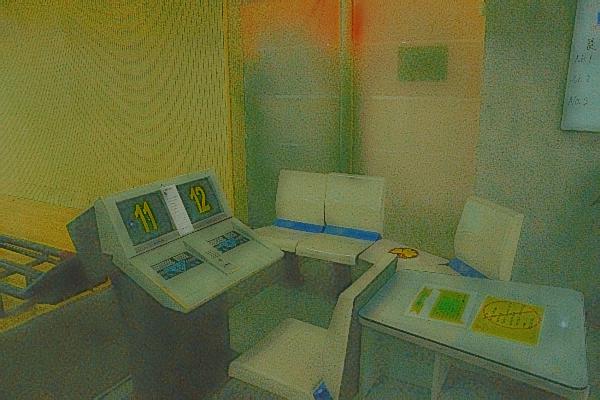}}
   \vspace{1.5pt}
   \centerline{RetinexNet}
   
 \end{minipage}
 \begin{minipage}[t]{0.137\linewidth}
  \centerline{\includegraphics[width=\textwidth]{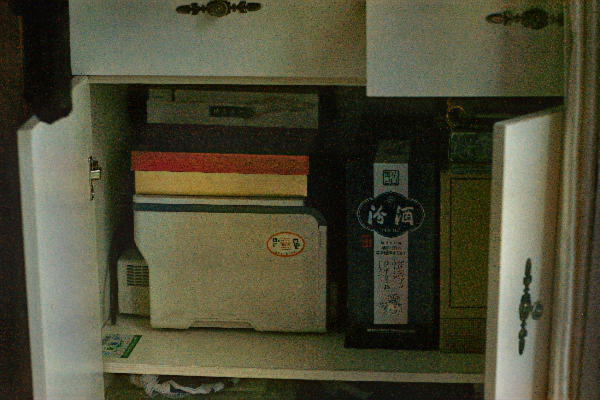}}
  \vspace{1.5pt}
  \centerline{\includegraphics[width=\textwidth]{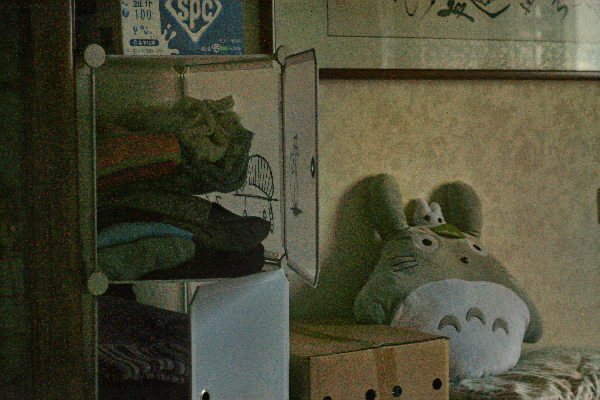}}
  \vspace{1.5pt}
  \centerline{\includegraphics[width=\textwidth]{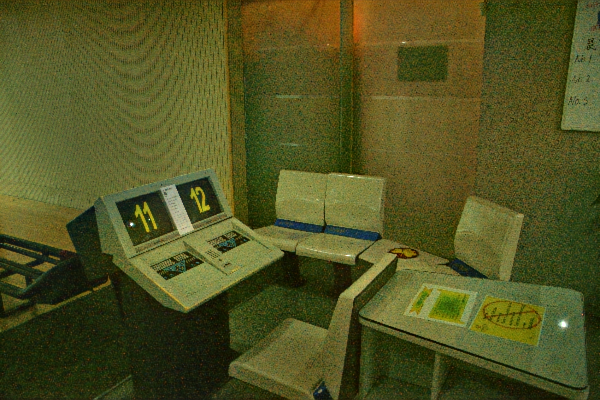}}
   \vspace{1.5pt}
   \centerline{EnlightenGAN}
 \end{minipage}
 \begin{minipage}[t]{0.137\linewidth}
  \centerline{\includegraphics[width=\textwidth]{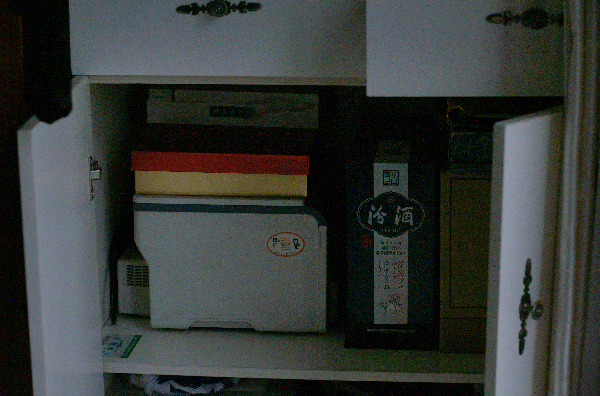}}
  \vspace{1.5pt}
  \centerline{\includegraphics[width=\textwidth]{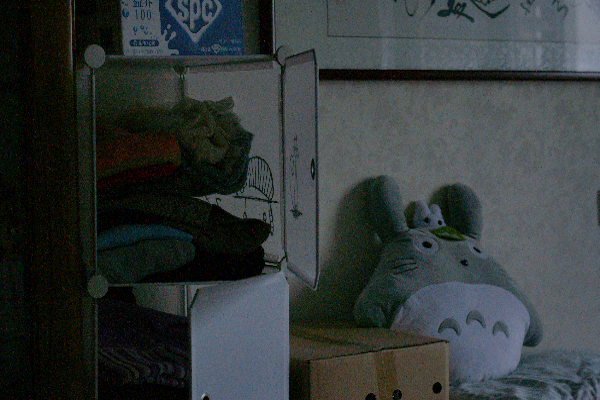}}
  \vspace{1.5pt}
  \centerline{\includegraphics[width=\textwidth]{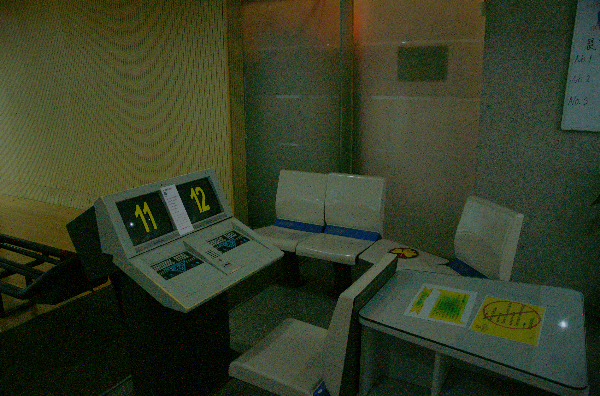}}
   \vspace{1.5pt}
   
   \centerline{Zero-DCE++}
   
 \end{minipage}
 \begin{minipage}[t]{0.137\linewidth}
  \centerline{\includegraphics[width=\textwidth]{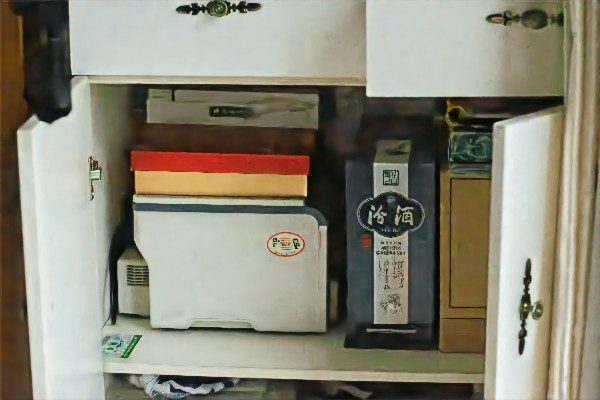}}
  \vspace{1.5pt}
  \centerline{\includegraphics[width=\textwidth]{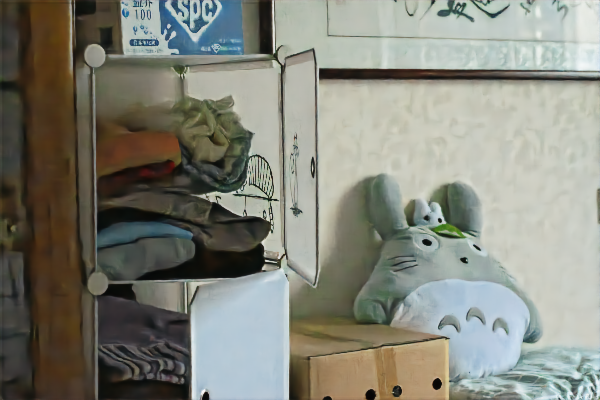}}
  \vspace{1.5pt}
   \centerline{\includegraphics[width=\textwidth]{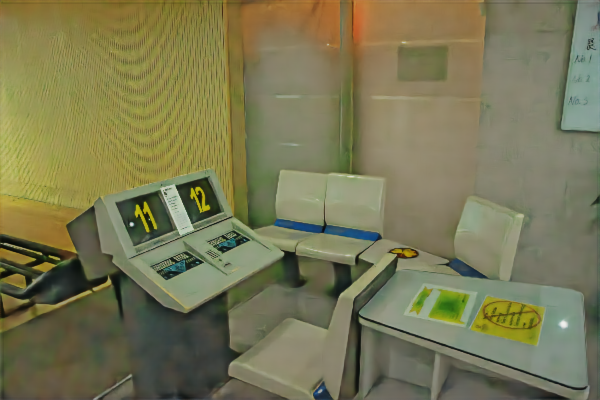}}
   \vspace{1.5pt}
   \centerline{KinD++}
 \end{minipage}
 \begin{minipage}[t]{0.137\linewidth}
  \centerline{\includegraphics[width=\textwidth]{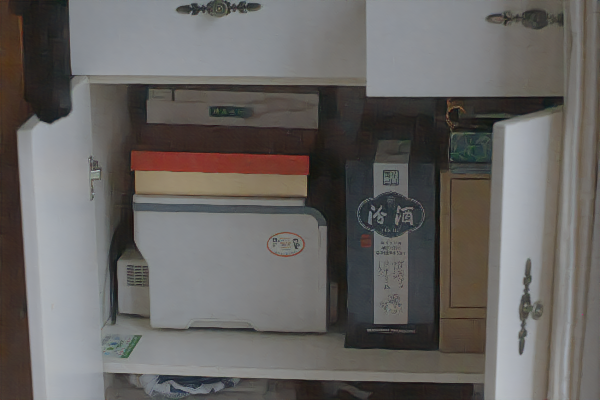}}
  \vspace{1.5pt}
  \centerline{\includegraphics[width=\textwidth]{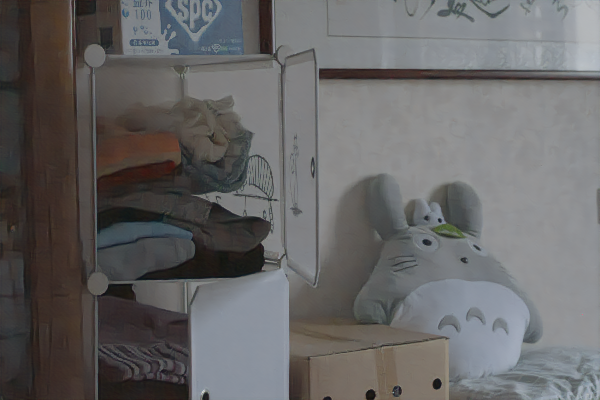}}
  \vspace{1.5pt}
\centerline{\includegraphics[width=\textwidth]{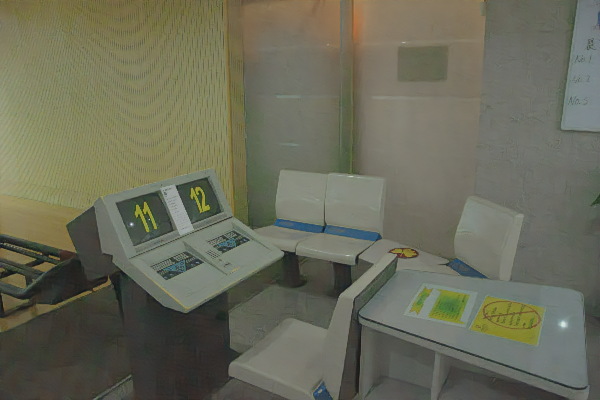}}
  \vspace{1.5pt}
  \centerline{URetinex-Net}
 \end{minipage}
 \begin{minipage}[t]{0.137\linewidth}
  \centerline{\includegraphics[width=\textwidth]{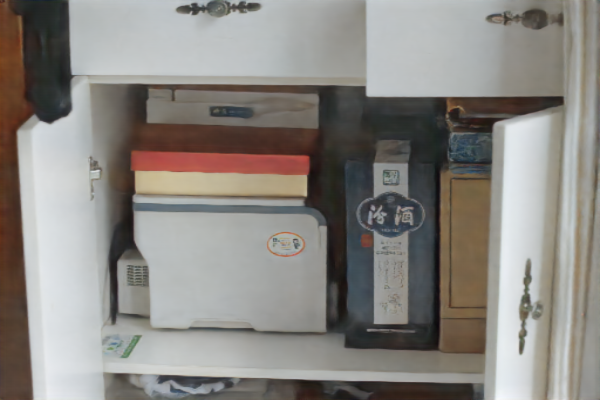}}
  \vspace{1.5pt}
  \centerline{\includegraphics[width=\textwidth]{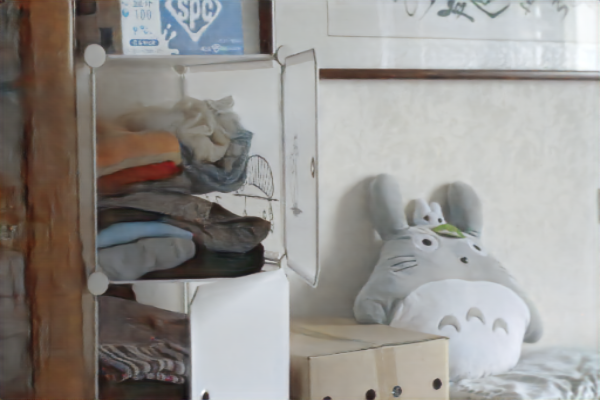}}
  \vspace{1.5pt}
  \centerline{\includegraphics[width=\textwidth]{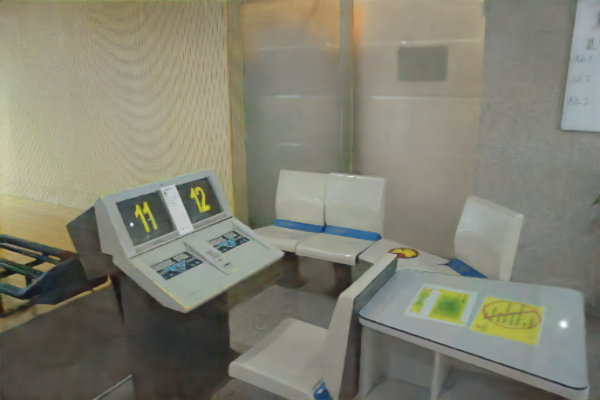}}
  \vspace{1.5pt}
  \centerline{ClassLIE (Ours)}
 \end{minipage} 
\caption{A comparison of the visual results with the SoTA methods on the LOL v1 dataset. Our method demonstrates better edge preservation and illumination naturalness, particularly in the magnified regions.}
 \label{LOL-testing}
\end{center}
\vspace{-16pt}
\end{figure*}

\section{Experiments}
In this section, we demonstrate the capability of our proposed ClassLIE model quantitatively and qualitatively on the widely-adopted low-light benchmarks compared with other SoTA LLE methods.
\vspace{-10pt}
\subsection{Experimental Settings}
\noindent\textbf{Implementation Details.}~We conduct experiments with PyTorch\cite{paszke2019pytorch} on an NVIDIA GeForce RTX 4090 GPU. All training and test images are resized into $ 256 \times 256$ and the patch size is set to $ 32 \times 32 $. Two network training phases are implemented in ClassLIE for the SIC module and the FLF module. In the SIC module, the batch size and the number of epochs are set to 256 and 400, respectively. The learning utilizes the cosine annealing algorithm, with the initial learning rate to be $2\times 10^{-4}$, a cycle of 5 epochs, and a minimum value of 0.
In the FLF module, the batch size is 2, the number of epochs is $2\times 10^{4}$, and the learning rate is decreased by 0.9 every 3000 epochs. For both parts, we employed the Adam optimizer, with the corresponding beta1 and beta2 set as (0.9, 0.999). During the training process, we test the model inference capability every 100 epochs and compare it with the saved model for optimal preservation.

\noindent\textbf{Datasets.}~We train our ClassLIE using LOL real-world data~\cite{wei2018deep}, which includes 485 paired images. We use the datasets of LOL~\cite{wei2018deep} (15 paired images), LIME~\cite{guo2016lime} (10 images), MEF \cite{ma2015perceptual} (17 images), NPE \cite{wang2013naturalness} (85 images) and DICM \cite{lee2013contrast} (64 images) for test.

\noindent\textbf{Running Time.}~The overall training duration of the network on a single 4090 GPU is approximately 82.6 hours, and the time to train and test a single image is around 1.69 seconds and 1.03 seconds respectively. 

\noindent\textbf{Evaluation Metrics.}~For the LOL test dataset (paired images), we use two metrics, PSNR \cite{hore2010image} and SSIM \cite{wang2004image}
for evaluation. A higher PSNR score means better performance in suppressing artifacts and preserving color information. A higher SSIM score denotes a better ability to preserve structure details. 
For LIME, MEF, NPE, and DICM datasets (without paired images), NIQE~\cite{mittal2012making} is used. The smaller the NIQE, the better the naturalness of the enhanced results is. 



\noindent\textbf{Compared Methods.}~We compare our ClassLIE with four groups of SoTA learning-based LIE methods:({I}) methods that incorporate the Retinex theory, \eg,~RetinexNet \cite{wei2018deep}, KinD \cite{zhang2019kindling}, KinD++ \cite{zhang2021beyond} and URetinex-Net \cite{wu2022uretinex}; ({II}) methods that mainly consider the prior information, \eg,~DCC-Net\cite{zhang2022deep} and SMG \cite{SMG}; ({III}) unsupervised methods, \eg,~EnlightenGAN \cite{jiang2021enlightengan}; ({IV}) zero-shot learning methods, \eg~Zero-DCE \cite{guo2020zero} and Zero-DCE++ \cite{li2021learning}.
\vspace{-12pt}
\subsection{Quantitative and Qualitative Results}
 We conducted the final qualitative experiments by repeating the training 12 times on different GPUs, including NVIDIA RTX 6000 Ada, 4090, and 3090 Ti, while maintaining the same network settings. The ClassLIE results presented in Tab.~\ref{LOL-quan} and Tab.~\ref{unpaired-quan} represent the average values obtained from these trials. We have also displayed the standard deviation in these tables to illustrate the consistency across different hardware configurations.
\subsubsection{Paired Data (LOL).}
The quantitative results are shown in Tab.~\ref{LOL-quan}. Our method which considers both the structure and illumination priors, outperforms all the other methods in terms of PSNR (25.74 dB) and SSIM (0.92). Specifically, our method surpasses the SoTA structural guided method SMG by more than 0.12 dB in PSNR and 0.01 in SSIM, respectively. And we achieved more than 3 in PSNR and 0.1 in SSIM compared with the second-best method.
The qualitative results are shown in Fig.~\ref{LOL-testing}. Our method better suppresses the unbalanced exposure and noise while recovering the detailed structural and illumination information. For other methods, \eg,~RetinexNet, EnlightenGAN, and ZeroDCE++, the structural details are missing, and exposure imbalance also appears in the enhanced images. Moreover, the results of RetinexNet look blurry. The results of KinD++ and URetinex-Net preserve most of the structural details. However, the results of KinD++ seem to be over-enhanced (\eg,~have color distortion and are brighter compared with the ground truth), this might be due to the neglecting of the local or color information in their network design). URetinex-Net has the inaccurate color for the local regions (\eg,~the blue book has some green color) and is slightly darker than our results.

\subsubsection{Unpaired Data (LIME, MEF, NPE, DICM).}
\begin{figure*}[t!]
 \begin{center}
 \begin{minipage}[t]{0.137\linewidth}
  \centerline{\includegraphics[width=\textwidth]{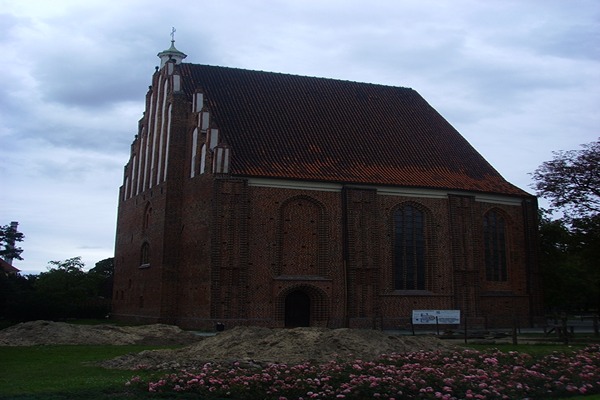}}
  \vspace{1.5pt}
  \centerline{\includegraphics[width=\textwidth]{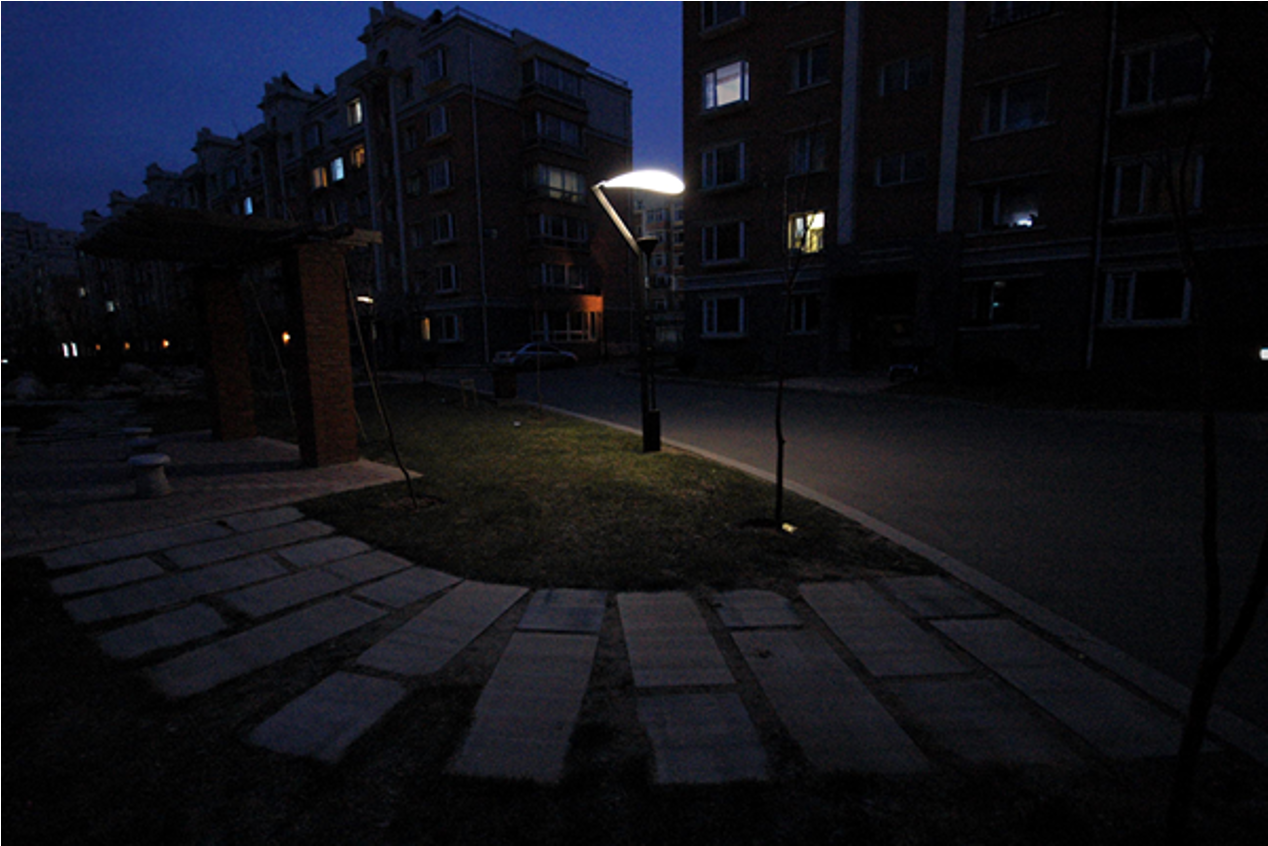}}
  \vspace{1.5pt}
  \centerline{\includegraphics[width=\textwidth]{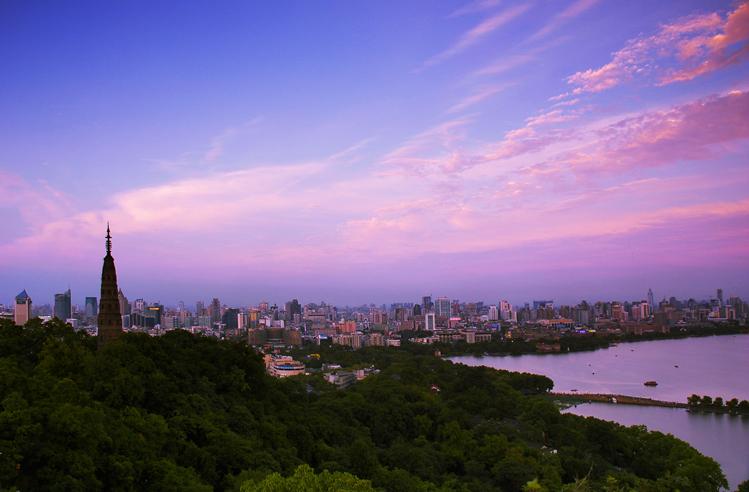}}
  \vspace{1.5pt}

  \centerline{Input}
 \end{minipage}
 \begin{minipage}[t]{0.137\linewidth}
  \centerline{\includegraphics[width=\textwidth]{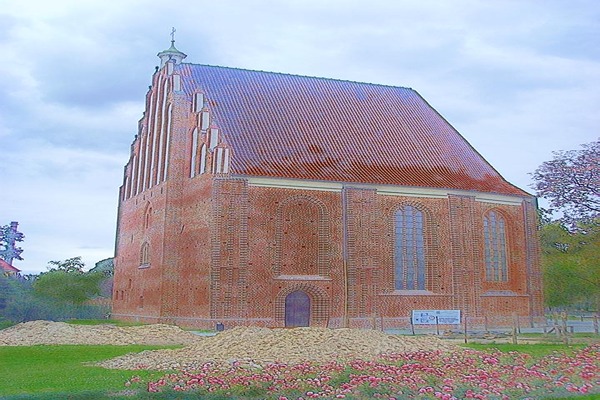}}
  \vspace{1.5pt}
\centerline{\includegraphics[width=\textwidth]{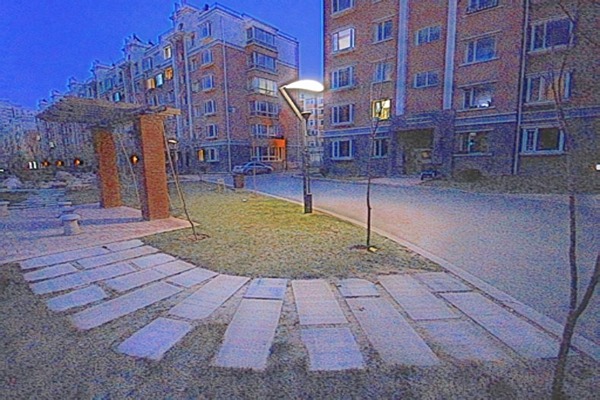}}
\vspace{1.5pt}
 \centerline{\includegraphics[width=\textwidth]{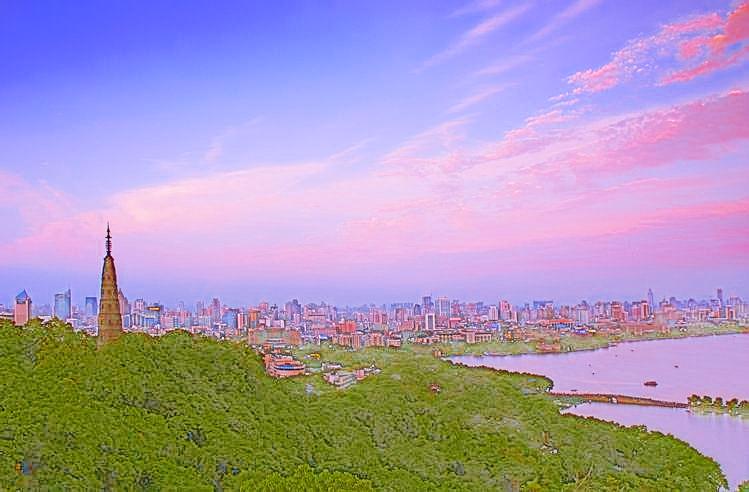}}
  \vspace{1.5pt}

  \centerline{RetinexNet}
 \end{minipage}
 \begin{minipage}[t]{0.137\linewidth}
  \centerline{\includegraphics[width=\textwidth]{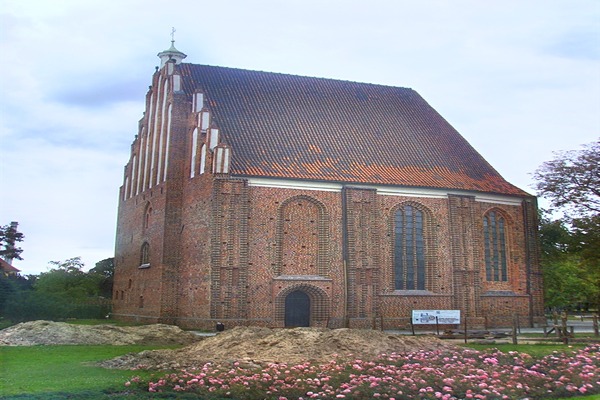}}
  \vspace{1.5pt}
  \centerline{\includegraphics[width=\textwidth]{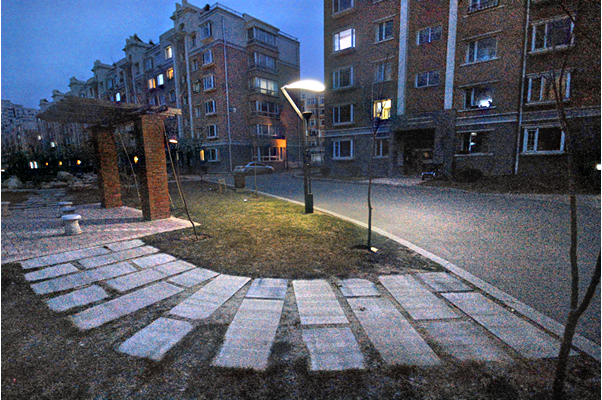}}
  \vspace{1.5pt}
  \centerline{\includegraphics[width=\textwidth]{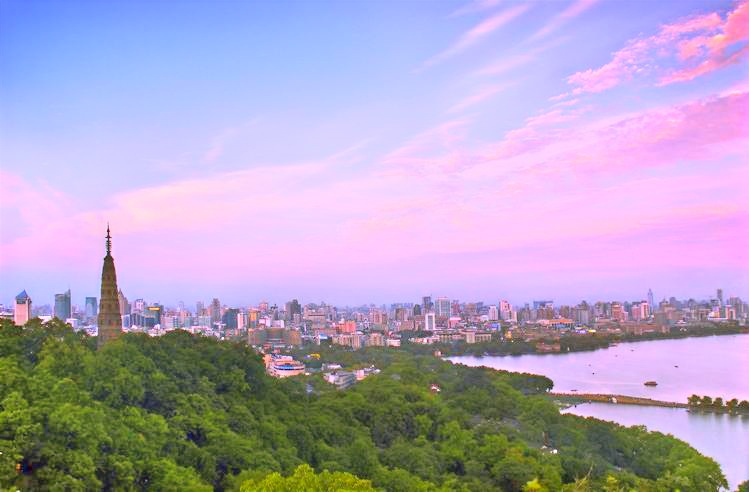}}
  \vspace{1.5pt}
 
  \centerline{EnlightenGAN}
 \end{minipage}
 \begin{minipage}[t]{0.137\linewidth}
  \centerline{\includegraphics[width=\textwidth]{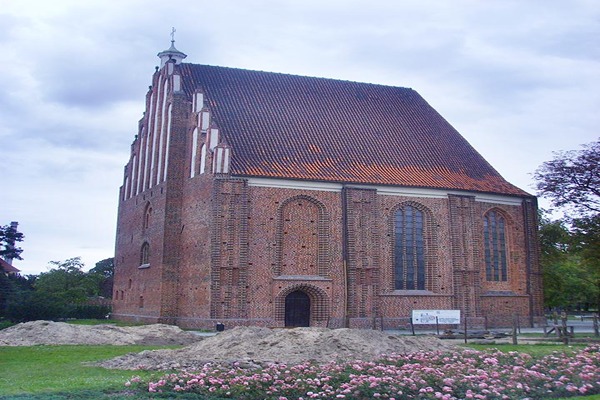}}
  \vspace{1.5pt}
  \centerline{\includegraphics[width=\textwidth]{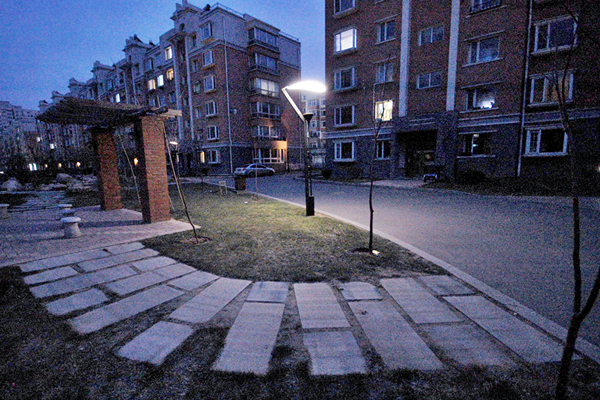}}
  \vspace{1.5pt}
 \centerline{\includegraphics[width=\textwidth]{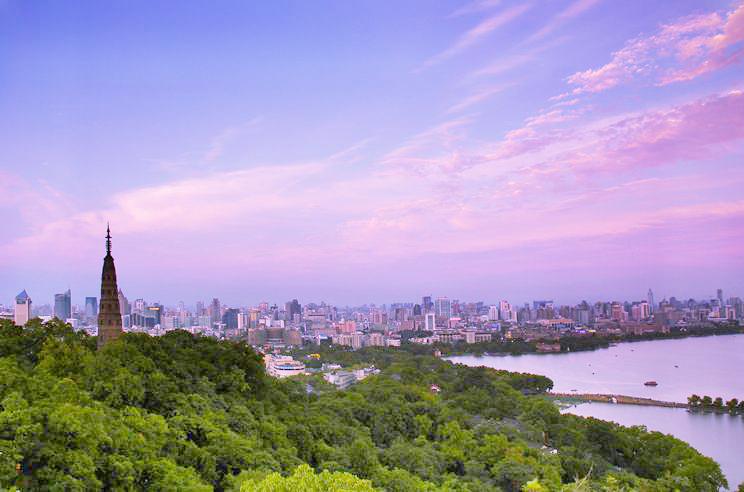}}
  \vspace{1.5pt}

  \centerline{Zero-DCE++}
 \end{minipage} 
 \begin{minipage}[t]{0.137\linewidth}
  \centerline{\includegraphics[width=\textwidth]{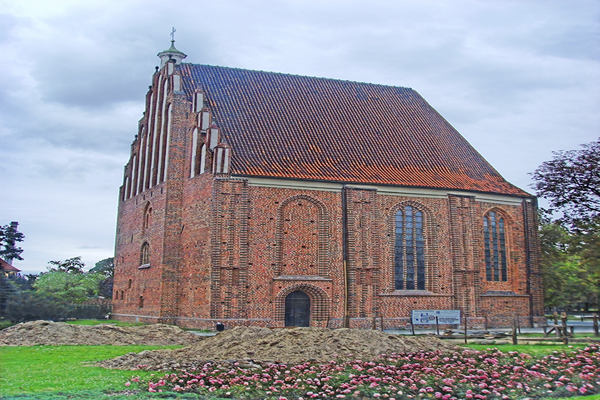}}
  \vspace{1.5pt}
  \centerline{\includegraphics[width=\textwidth]{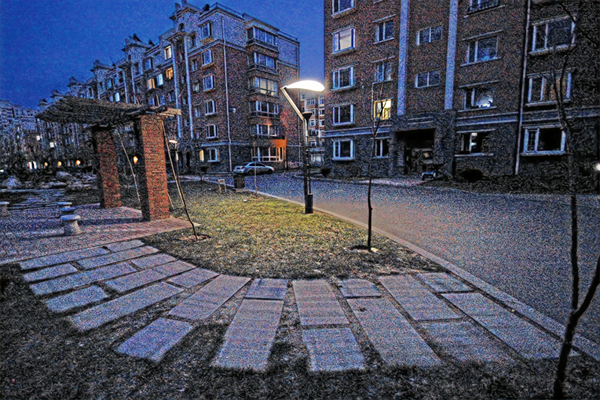}}
  \vspace{1.5pt}
  \centerline{\includegraphics[width=\textwidth]{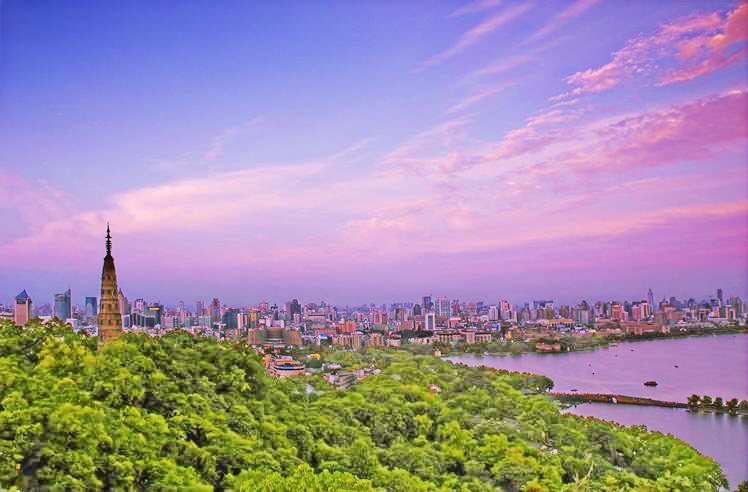}}
  \vspace{1.5pt}
 
  \centerline{KinD++}
 \end{minipage}
 \begin{minipage}[t]{0.137\linewidth}
  \centerline{\includegraphics[width=\textwidth]{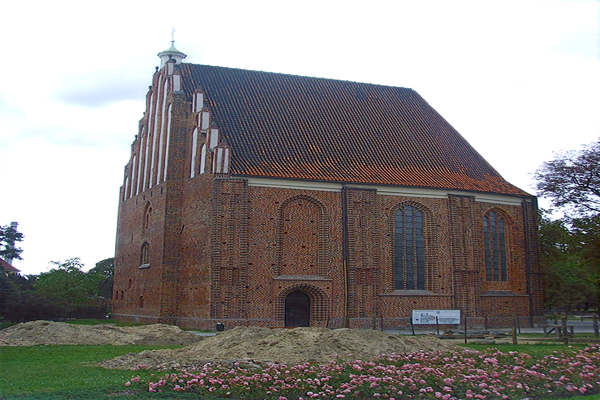}}
  \vspace{1.5pt}
  \centerline{\includegraphics[width=\textwidth]{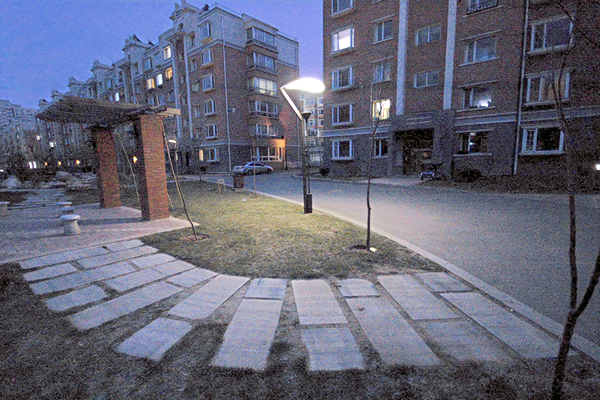}}
  \vspace{1.5pt}
  \centerline{\includegraphics[width=\textwidth]{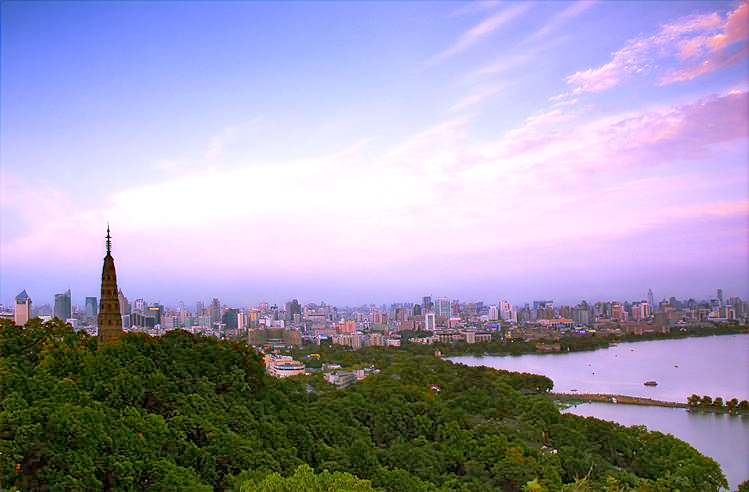}}
  \vspace{1.5pt}

  \centerline{URetinex-Net}
 \end{minipage}
 \begin{minipage}[t]{0.137\linewidth}
  \centerline{\includegraphics[width=\textwidth]{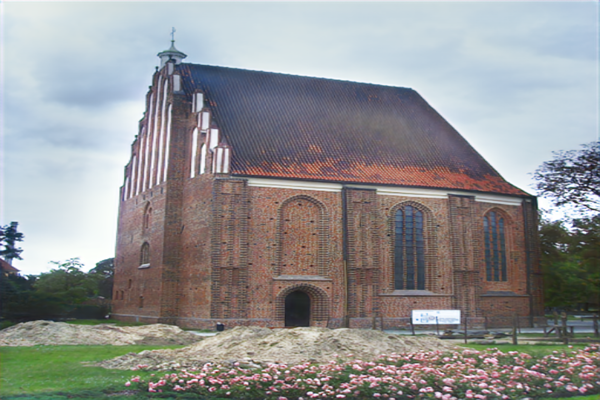}}
  \vspace{1.5pt}
  \centerline{\includegraphics[width=\textwidth]{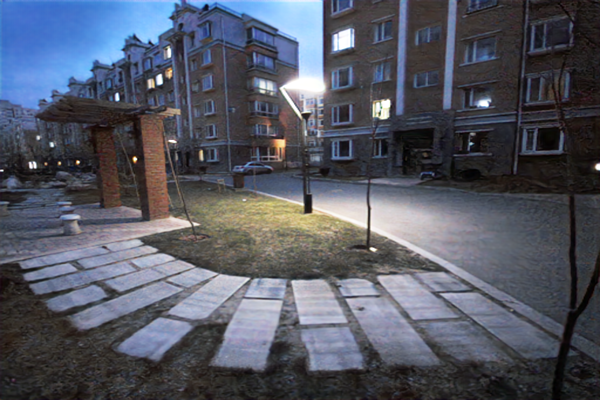}}
  \vspace{1.5pt}
  \centerline{\includegraphics[width=\textwidth]{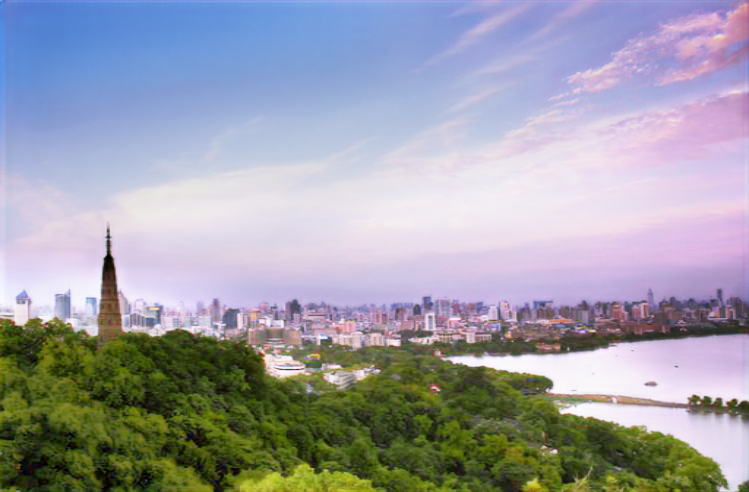}}
  \vspace{1.5pt}
  
  \centerline{ClassLIE(Ours)}
 \end{minipage}
 \caption{A comparison of the visual results with the SoTA methods on DICM (first row), LIME (second row), NPE (third row). On the real-world dataset, our enhancement results still exhibit superior structural and illumination information.}
 \label{unpaired}
\end{center}
\end{figure*}

Tab.~\ref{unpaired-quan} shows the quantitative results in terms of NIQE. In general, our ClassLIE obtains the best results for all four unpaired datasets. It is worth mentioning that our result under the MEF dataset is lower by a significant value of 1.25, compared with the existing sota performance. The visual results are depicted in Fig. \ref{unpaired}. We observe that: 1) RetinexNet, EnlightenGAN and KinD++ produce unnatural results with obvious color distortion and noise. 2) URetinex-Net suffers from local over-exposure problems (\eg,~The sky is overexposed). 3) Zero-DCE++ achieves better visual quality; however, the results are noisy. By contrast, our results look smoother, especially for images with complex structures and illumination distribution.
\vspace{-20pt}

\begin{table}[t!]
\caption{Metric Comparison on the LOL Dataset in terms of PSNR and SSIM.}
\centering
\setlength{\tabcolsep}{8pt}
\begin{threeparttable}
\begin{tabular}{l l l c c}
\toprule
Method             & SP &IP & PSNR $\uparrow$ & SSIM $\uparrow$ \\ \hline
RetinexNet (BMVC’18)    & $\times$ & $\times$ & 16.82    & 0.43      \\
KinD (MM’19)       & $\checkmark$ & $\times$ & 20.42    & 0.82       \\
Zero-DCE (CVPR’20)   & $\times$ & $\times $ & 14.86          & 0.54      \\
EnlightenGAN (TIP’21) & $\times$ & $\checkmark$  & 17.48          & 0.65       \\
ZeroDCE++ (TPAMI’21)  & $\times$  & $\times$ & 16.11          & 0.53    \\
KinD++ (IJCV’21)      & $\checkmark$ & $\times$ & 21.30          & 0.82     \\
LLFlow (AAAI'22)  & $\times$ & $\checkmark$ & 18.88          & 0.69    \\
URetinex-Net (CVPR'22) & $\checkmark$ & $\times$ & 21.33          & 0.83    \\
DCC-Net (CVPR'22)  & $\checkmark$ & $\times$ & {22.72}          & 0.81    \\
SMG (CVPR'23)  & $\checkmark$ & $\times$ & \textcolor{blue}{25.62}          & \textcolor{blue} {0.91}    \\
ClassLIE (Ours)    & $\checkmark$ & $\checkmark$ & \textcolor{red}{25.74±0.07}       & \textcolor{red}{0.92±0.004} \\ 
\bottomrule
\end{tabular}

\begin{tablenotes}
\small
\item {*\textbf{SP} in the table indicates structure prior information and \textbf{IP} indicates illumination prior information.}

\end{tablenotes}
\end{threeparttable}
\label{LOL-quan}
\end{table}


\begin{table}[t!]
\renewcommand{\tabcolsep}{7pt}
\centering
\caption{Metric Comparison on LIME, MEF, NPE, DICM in terms of NIQE}
\begin{tabular}{l c c c c}
\toprule
Method  & LIME &  MEF & NPE  & DICM  \\ \hline
RetinexNet    & 5.75  & 4.93  & 4.95  &4.33  \\ 
KinD          & \textcolor{blue}{4.42}  & \textcolor{blue}{4.45} & 3.92 & 3.95   \\ 
Zero-DCE      & 5.82  & 4.93   & 4.53   &4.58 \\ 
EnlightenGAN  & 4.59  & 4.70   &3.99   &4.06 \\  
Zero-DCE++    & 5.66  & 5.10   &4.74.  &4.89 \\  
KinD++        & 4.90 & 4.55    &\textcolor{blue}{3.91}   &\textcolor{blue}{3.89} \\  
URetinex-Net      & \textcolor{blue}{4.42} & 4.59 &4.13 &4.54 \\ 
{ClassLIE}      & {\textcolor{red}{{3.23±0.1}}} & {\textcolor{red}{{3.20±0.1}}}   &{\textcolor{red}{{3.43±0.03}}} & {\textcolor{red}{{3.64±0.06}}}  \\ \bottomrule
\end{tabular}
\label{unpaired-quan}
\end{table}

\subsection{User Study}
Due to the lack of a clearly defined `ground truth' in LIE tasks, we also conducted a user study based on human subjective visual perception to determine the naturalness of enhancement results and the recognizability of details. Since our network mainly considers the restoration effect of illumination information and structural information, we have set two indicators named as naturalness index (NI) and clarity index (CI), which adopt a 5-point scoring system. To prevent bias from human cognition, we invited 100 volunteers with different knowledge backgrounds in the field of computer vision to carry out this user study with their NI and CI ratings for each image. We selected a total of 10 raw inputs from the LOL, LIME, MEF, NPE, and DICM test sets and applied the pre-trained models of RetinexNet, Kind, Zero-DCE, EnlightenGAN, Zero-DCE++, KinD, KinD++, URetinex, and our ClassLIE for testing. Finally, we summarized the obtained 80 results from the survey questionnaire and calculated the average NI and CI for each network, as shown in Tab.~\ref{userstudy}.

\begin{table}[t!]
\renewcommand{\tabcolsep}{21pt}
\centering
\caption {Average naturalness index (NI) and clarity index (CI) collected from user study }
\begin{tabular}{l c c }
\toprule
Method  & NI & CI  \\ \hline
RetinexNet    & 2.19  & 2.53    \\ 
Zero-DCE      & 2.96  & 3.32  \\ 
EnlightenGAN  & 2.59  & 3.15   \\  
Zero-DCE++    &  \textcolor{blue}{3.44} & 3.48   \\
KinD   &  2.74 & 2.92  \\ 
KinD++        & 2.89 & 3.07  \\  
URetinex-Net      & 3.11 & \textcolor{blue}{3.59} \\ 
{ClassLIE}      & {\textcolor{red}{{3.46}}} & {\textcolor{red}{{3.81}}}    \\ \bottomrule
\end{tabular}
\label{userstudy}
\end{table}

From Tab.~\ref{userstudy}, it is witnessed that our proposed ClassLIE has achieved the best score in user ratings. It indicates that the vast majority of users believe that our enhanced results have the most natural and clearest visual effects. Especially in terms of CI which is related to structural information, our score is much higher than other networks (0.24 higher than the second highest). At the same time, it can be observed that unsupervised networks such as the Zero-DCE series generally perform well in user study.

\subsection{Ablation Study and Analysis}
\noindent\textbf{Effectiveness of Baseline.} We validate the contribution of different baselines in terms of PSNR and SSIM in Tab.~\ref{baseline}. The enhancement results produced by simply using U-net, DnCNN net, SwinIR, and the combination of SwinIR and the DnCNN are presented in the table. We can observe that the combination of the SwinIR and DnCNN achieves the best results. Therefore, in our design, we use the modified SwinIR and DnCNN as our backbone architecture.

\begin{table}[t!]
\renewcommand{\tabcolsep}{15pt}
\centering
\caption{The Effectiveness of Different Backbones on the LOL Dataset.
}
\begin{tabular}{l l l}
\toprule
Method  &  PSNR $\uparrow$ &  SSIM $\uparrow$   \\ \hline
U-Net           & 16.83  & 0.66     \\ 
DnCNN          & 16.97  & 0.66   \\ 
SwinIR         & 19.26  & 0.78    \\ 
SwinIR+DnCNN   & \textcolor{red}{19.97}  & \textcolor{red}{0.83}      \\  
\bottomrule
\end{tabular}

\label{baseline}
\end{table}

\noindent\textbf{Effectiveness of SIC module.}
Tab.~\ref{Classification} presents the ablation results in terms of PSNR and SSIM of our SIC module on the LOL dataset. When we experiment without SIC, the CAFL simply uses 5 layers of DnCNN architecture but removes the batch normalization, which is consistent throughout our experiment. From the table, “w/o SIC” means enhancement results without the SIC module, which is also our baseline performance; the “SIC-S(w/o LD)” means the SIC calculates only structure information on original inputs, removing layer decomposition; while the “SIC-S+I(w/o LD)” means the SIC is according to both the structure information and the illumination information on raw RGB input, removing layer decomposition; and “SIC-S(w/ LD)” means the classification is applied based on structure information on decomposed reflectance map, while “SIC-S+I(w/ LD)” denotes an integral SIC module which is also adopted by ClassLIE eventually. We can see a noticeable performance decline without layer decomposition or an illumination map in SIC, which demonstrates the rationality and validity of the proposed SIC module based on both the structure and illumination information.
\begin{table}[t!]
\renewcommand{\tabcolsep}{17pt}
\centering
\caption{Ablation Study of the Contribution of SIC-Net in ClassLIE on LOL dataset in terms of PSNR and SSIM}
\begin{threeparttable}

\begin{tabular}{l l l}
\toprule
Method  & PSNR $\uparrow$ &  SSIM $\uparrow$    \\ \hline
w/o SIC          & 17.33  & 0.80    \\ 
SIC-S(w/o LD) & 23.10  & 0.81   \\ 
SIC-S+I(w/o LD)  & 24.61  & 0.84  \\
SIC-S(w/ LD) & 24.72  & 0.88    \\ 
SIC-S+I(w/ LD)   & \textcolor{red}{25.74}  & \textcolor{red}{0.92}     \\  
\bottomrule
\end{tabular}
\vspace{1.5pt}
\begin{tablenotes}
\small
\item {*\textbf{S} in the table indicates structure information and \textbf{I} indicates illumination information. \textbf{LD} indicates layer decomposition network.}
\end{tablenotes}
\end{threeparttable}

\label{Classification}
\end{table}

\noindent\textbf{Effectiveness of numbers of Dformer blocks.}~Tab.~\ref{layer} shows the results by varying the number of Dformer blocks. It can be seen that adding extra Dformer blocks brings marginal gains when the number is larger than two while bringing additional computational costs. Therefore, we chose two Dformer blocks in the final framework. This effect is mostly caused by the model being too deep due to an increase in Dformer blocks, which results in poor generalization and overfitting problems. In our experiment, we found the performance of our method, when using more than two Dformer blocks. Additionally, too many Dformer blocks make the model more difficult to optimize, since each Dformer Block contains two complex branches (PRL and CAFL), making it challenging to find the global minimum within a limited number of iterations.


\begin{table}[t!]\centering
\begin{center}
\renewcommand{\tabcolsep}{15pt}
\caption{Ablation Study of the Numbers of Dformer Blocks. 
}
\centering
\begin{tabular}{c l l}
\toprule
Dformer Blocks  & PSNR $\uparrow$ &  SSIM $\uparrow$    \\ \hline
1 & 23.27  & 0.79  \\ 
2 & 25.74  & 0.92  \\ 
6 & 25.75  & 0.92   \\
\bottomrule
\end{tabular}
\label{layer}
\end{center}
\end{table}

\begin{figure}[t]
\begin{center}
\begin{minipage}[t]{0.24\linewidth}
\centerline{\includegraphics[width=\textwidth]{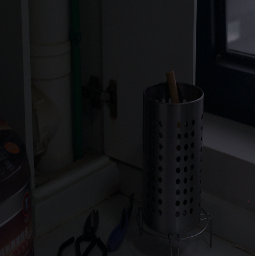}}
\centerline{Input}
\end{minipage}
\begin{minipage}[t]{0.24\linewidth}
\centerline{\includegraphics[width=\textwidth]{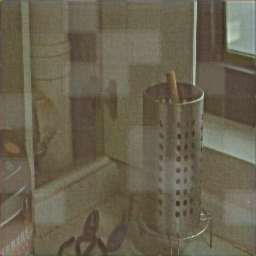}}
\centerline{w/o fusion}
\end{minipage}
\begin{minipage}[t]{0.24\linewidth}
\centerline{\includegraphics[width=\textwidth]{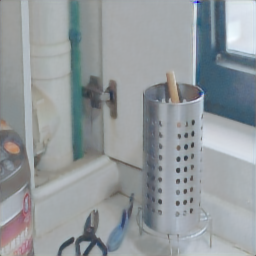}}
\centerline{w/ fusion}
\end{minipage}
\begin{minipage}[t]{0.24\linewidth}
\centerline{\includegraphics[width=\textwidth]{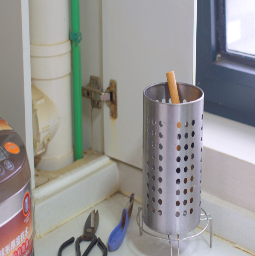}}
\centerline{GT}
\end{minipage}
 \caption{Visual illustration of our baseline method test on the LOL dataset. The first column is the input low-light images; the second column is the visual results without feature fusion, where there are ``checkerboard artifacts"; the 3rd column shows the visual results with feature fusion; the last column is the reference ground truth image.}
 \label{FF}
 \end{center}
\end{figure}

\noindent\textbf{Effectiveness of feature fusion in FLF.} As in the FLF module, we enhance the low-light images based on the patches with one branch through the Transform architecture and another branch through the CNN-based architecture. We simply make a superposition of the two branch results when our proposed feature fusion is not used.
Fusing the two features properly is a significant challenge. When features are not fused in the right way, problems such as the ``black shadows artifacts" or the ``checkerboard artifacts" will worsen the PSNR and SSIM values~(See~Fig.~\ref{FF}). Tab.~\ref{Blending} also demonstrates the effectiveness of our feature fusion block which models the dependencies between the channels from the two features. 

\begin{table}[t!]\centering
\renewcommand{\tabcolsep}{14pt}
\caption{The Effectiveness of Feature Fusion on the LOL Dataset.}
\centering
\begin{tabular}{l l l}
\toprule
Method  & PSNR $\uparrow$ &  SSIM $\uparrow$    \\ \hline
w/o Feature Fusion    & 12.40  & 0.33  \\ 
w/ Feature Fusion   & \textcolor{red}{25.74}  & \textcolor{red}{0.92}     \\  
\bottomrule
\end{tabular}
\label{Blending}
\end{table}

\noindent\textbf{Color Consistency.}~To avoid color bias and preserve color consistency, we perform three ablation studies to verify the effectiveness of color consistency: existing RGB color losses from ZeroDCE \cite{guo2020zero} (w/ $L_{c-DCE}$) and color map from LLflow \cite{wang2022low} (w/ $L_{c-Flow}$) with our reflectance color map (w/ $L_{c-ClassLIE}$). 
We explore their contributions to the performance of ClassLIE in terms of PSNR and SSIM. 
Tab.~\ref{color} demonstrates that using the reﬂectance for color enhancement results in large performance gains (PSNR + 1.22 dB, SSIM + 0.05) compared with the second best with $L_{c-Flow}$.

\begin{table}[t!]
\caption{Ablation Study of the Color Consistency in ClassLIE on LOL dataset.}
\renewcommand{\tabcolsep}{17pt}
\centering
\begin{tabular}{l l l}
\toprule
Method  & PSNR $\uparrow$ &  SSIM $\uparrow$\\ \hline
w/ $L_{c-DCE}$    & 21.14  & 0.82   \\ 
w/ $L_{c-Flow}$ & 22.07  & 0.83    \\ 
w/ $L_{c-ClassLIE}$   & \textcolor{red}{25.74} & \textcolor{red}{0.92} \\  
\bottomrule
\end{tabular}
\vspace{1.5pt}
\label{color}
\end{table}

\begin{figure}[t]
 \begin{center}
 \begin{minipage}[t]{0.24\linewidth}
   \centerline{\includegraphics[width=\textwidth]{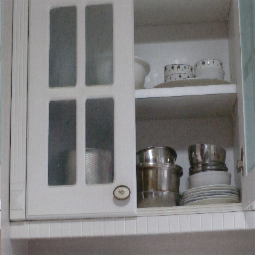}}
   \vspace{1pt}
   \centerline{$L_{c-DCE}$}
 \end{minipage}
 \begin{minipage}[t]{0.24\linewidth}
   \centerline{\includegraphics[width=\textwidth]{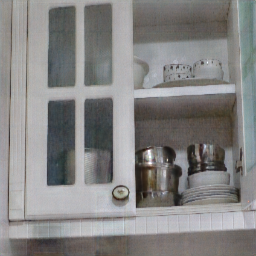}}
   \vspace{1pt}
   \centerline{$L_{c-Flow}$}
 \end{minipage}
  \begin{minipage}[t]{0.24\linewidth}
  \centerline{\includegraphics[width=\textwidth]{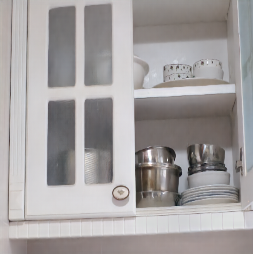}}
  \vspace{1pt}
  \centerline{$L_{c-ClassLIE}$}
 \end{minipage}
  \begin{minipage}[t]{0.24\linewidth}
   \centerline{\includegraphics[width=\textwidth]{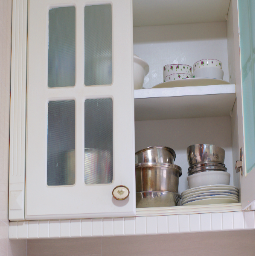}}
   \vspace{1pt}
   \centerline{GT}
 \end{minipage}
 \vspace{-6pt}
 \caption{Visual result of applying different color consistency approaches. The first column is the visual result of applying RGB color losses from ZeroDCE; the second column is the visual results of using LLflow color map; the third column is the visual results of using our reflectance color map; the last column shows the ground truth. It witnessed minimal color distortion by using our proposed color map}
 \label{CC}
\end{center}
\vspace{-5pt}
\end{figure}

\begin{table}[t!]\centering
\begin{center}
\caption{The Results of the Different Number of Classification Levels in SIC Module on the LOL Dataset.}
\renewcommand{\tabcolsep}{12pt}
\begin{tabular}{c l l l}
\toprule
Num.  &  PSNR $\uparrow$ &  SSIM $\uparrow$   &  Params. $\downarrow$ \\ \hline
2 & 20.33  & 0.79 & 12.24\small~M \\ 
3 & 25.74  & 0.92 & 16.68\small~M  \\ 
6 & 25.74  & 0.91 & 22.01\small~M \\ 
10 & 25.81  & 0.92  & 27.31\small~M\\ 
\bottomrule
\end{tabular}
\label{levels}
\vspace{-18pt}
\end{center}
\end{table}

\noindent\textbf{Different classification of difficulty levels.}~We divided the illumination levels into three categories based on the Retinex theory and the conventional definition of illumination categories (dark, average, and bright) \cite{ariffin2019illumination}. Meanwhile, we balance the performance and computation costs, as there shows negligible performance gain in illumination levels above 3 with an obvious increment of the parameters which introduces a lot of unnecessary computational cost to our method (See Tab.~\ref{levels}). In addition, the accuracy of the classification of difficulty levels has been clearly depicted in the plot of Fig.~\ref{class}, showing that each patch can be correctly classified into its correct category level.

\noindent\textbf{Computational Cost.}~ClassLIE has a relatively small number of parameters (Params:~16.68 M and FLOPs:~447.7 G) even than the related methods (Params:~0.84 M $\sim$ 115.63 M, FLOPs: 19.95 G $\sim$ 2087.35 G), as reported in \cite{cui2022illumination}. Actually, our ClassLIE better balances the model size and performance than the previous methods. Noted that we provided an ablation study for the number of transformer layers in Tab.~1 in the suppl. material. 
\vspace{-10pt}
\section{Conclusion}
This paper established a new framework, ClassLIE, that classifies and adaptively learns the structural and illumination information from the low-light images holistically and regionally. Our framework, which was built based on CNNs and transformers, was effective in achieving good enhancement performance. The experiments on five datasets demonstrated the superiority of our method, with good generalization ability to the unseen datasets. 

\textbf{Limitation and future work.}While our proposed method demonstrates commendable overall performance, it is essential to acknowledge certain limitations. The first challenge lies in the difficulty of acquiring a comprehensive understanding of sky-related features. Consequently, there are instances where a white hue effect emerges along the boundary between the sky and objects, accompanied by a decrease in sky color saturation. Addressing this limitation will involve a concerted effort towards refining patch relation learning and incorporating semantic knowledge about the sky. 
Secondly, the generality of our model is currently evaluated against a limited set of low-light images, which may not cover the full spectrum of degraded image types. This is a common challenge for the LIE community due to the difficult nature of capturing paired datasets. To enhance the model's robustness and versatility, future work will involve expanding the diversity of the dataset used for validation, particularly focusing on more complex real-world scenarios such as nighttime scenes and low-light images with backlit.
\vspace{-10pt}



\bibliographystyle{IEEEtran}

\vspace{-35pt}

\end{document}